\documentclass[preprint]{jmlr}
\usepackage{longtable}% for long tables

\usepackage{booktabs}
\usepackage{amsmath,amssymb,latexsym,amsfonts}
\usepackage{mathtools}
\usepackage{algorithm}
\usepackage{algorithmic}
\usepackage{bbm}
\usepackage{caption}

\usepackage{hyperref}
\usepackage{graphics}
\usepackage{graphicx}
\usepackage{color}
\usepackage{microtype}
\usepackage{xcolor}

\DeclareMathOperator*{\argmin}{arg\,min}

\jmlryear{2020}

\title[MetAL: Active Semi-Supervised Learning on Graphs via Meta Learning]{MetAL: Active Semi-Supervised Learning on Graphs via Meta Learning}

  \author{\Name{Kaushalya Madhawa} \Email{kaushalya@net.c.titech.ac.jp}\and
   \Name{Tsuyoshi Murata} \Email{murata@c.titech.ac.jp}\\
   \addr Department of Computer Science, School of Computing, \\
   Tokyo Institute of Technology, Tokyo, Japan}

\begin{document}

\maketitle

\begin{abstract}
The objective of active learning (AL) is to train classification models with less number of labeled instances by selecting only the most informative instances for labeling. The AL algorithms designed for other data types such as images and text do not perform well on graph-structured data. Although a few heuristics-based AL algorithms have been proposed for graphs, a principled approach is lacking. In this paper, we propose MetAL, an AL approach that selects unlabeled instances that directly improve the future performance of a classification model. For a semi-supervised learning problem, we formulate the AL task as a bilevel optimization problem. Based on recent work in meta-learning, we use the meta-gradients to approximate the impact of retraining the model with any unlabeled instance on the model performance. Using multiple graph datasets belonging to different domains, we demonstrate that MetAL efficiently outperforms existing state-of-the-art AL algorithms.
\end{abstract}

\section{Introduction}

The performance of a classification model depends on the quality and quantity of training data, often requiring a huge labeling effort. With ever-increasing amounts of data, active learning (AL) is gaining the attention of researchers as well as practitioners as a way to reduce the effort spent on labeling data instances. An AL algorithm selects a set of unlabeled instances based on an informative metric, gets their labels, and updates the labeled dataset. Then the classification model is retrained using the acquired labels. This process is repeated until a desirable level of performance (e.g. accuracy) is reached. 

In this paper, we consider the task of applying AL for semi-supervised problems. In a semi-supervised learning problem the learning algorithm can utilize all data instances including the unlabeled ones. Only the labels of unlabeled instances are not known. We evaluate our approach on classifying nodes on attributed graphs. Reducing the number of labeled nodes required in node classification can benefit a variety of practical applications such as in recommender systems \citep{pinsage18, rubens2015active} and text classification \citep{yao2019graph}.

An \textit{acquisition function} is used to evaluate the informativeness of an unlabeled instance. Since quantifying the \textit{informativeness} of an instance is not straightforward, a multitude of heuristics have been proposed in AL literature~\citep{settles2009active}. For example, \textit{uncertainty sampling} selects instances which the model is most uncertain about~\citep{houlsby2011bald}. The most common method is to select the unlabeled instance corresponding to the maximum entropy over the class probabilities predicted by the model.  However, such heuristics are not flexible to adapt to the distribution of data and can not exploit inherent characteristics of a given dataset. Often, the performance of heuristic active learners is not consistent across different datasets, sometimes worse than random selection of unlabeled instances.

Compared to applications of AL on image data, only a limited number of AL models have been developed for graph data. Previous work on applying AL on graph data~\citep{gu2012towards, bilgic2010networkdata, ji2012variance} depend on earlier classification models such as Gaussian random fields, in which the features of nodes are not being used. Therefore, selecting query nodes uniformly in random coupled with a  recent graph neural network (GNN) model can easily outperform such AL models. AL models that use recent GNN architectures~\citep{age2017cai, gao2018active} are limited and they rely on linear combinations of uncertainty and various heuristics such as node centrality measures. 

We overcome this problem by directly considering the performance of the classifier into the acquisition function in  semi-supervised learning problems. We build our work motivated by the framework of \textit{expected error reduction} (EER)~\citep{roy2001eer, guo2008discriminative, aodha2014hierarchical}, in which the objective is to query instances which would maximizes the expected performance gain. Original EER formulation is extremely time consuming and not practical to be used with neural network classifiers. We formulate this objective as a bilevel optimization problem. Based on recent advances in meta-learning \citep{finn2017maml}, we utilize meta-gradients to make this optimization efficient. \citet{zugner2018adversarial} propose using meta-gradients for modeling an adversarial attack on GNNs. Our motivation in using meta-gradients is the opposite, evaluating the importance of labeling each unlabeled instance. In \autoref{sec:results}, with empirical evidence, we show that MetAL significantly outperforms existing AL algorithms. 

%Additionally, we study the reasons contributing to the success of heuristics-based AL algorithms on commonly used citation networks. The citation datasets; CORA, CiteSeer \cite{sen2008cora}, and PubMed \cite{pubmed2012namata} are considered as the default datasets for benchmarking any node classification algorithm. We raise the question ``are these datasets suitable for benchmarking AL algorithms?". In answering this question, we find that AL algorithms tailored to be successful on citation datasets generalize poorly to other graph datasets.

Our  contributions are: 
\begin{enumerate}
	\item We introduce MetAL, a novel active learning algorithm based on the expected error reduction principle.
	\item We discuss the importance of performing exploration in AL and introduce a simple count-based exploration term.
	\item We demonstrate that our proposed algorithm MetAL can consistently outperform state-of-the-art AL algorithms on a variety of real world graphs.
\end{enumerate}

\section{Our Framework}

\subsection{Problem Setting}

In this paper, we apply AL for the multi-class node classification of a given undirected attributed graph $G$ of $N$ nodes. The graph $G$ consists of an adjacency matrix $A \in \{0, 1\}^{N \times N}$ and a node attribute matrix $X \in \mathbb{R}^{N \times F}$, where $F$ is the number of attributes. Labels of a small set of nodes $\mathcal{L}$ are given initially and labels of rest of the nodes $\mathcal{U}$ are unknown. A labeled node is assigned a label in $\{1,  2, \dots C\}$, where $C$ is the number of classes. The objective of a learner is to learn a function $f_\theta(x_i)$ which predicts the class label of a given test node $i \in \mathcal{U}$. This $f_{\theta}$ function can be any node classification algorithm. Graph neural networks \citep{kipf2017gcn, sgc2019} (GNN) are commonly used in the present day. Parameters $\theta$ of the model are estimated by minimizing a loss function, usually using a gradient-based optimization algorithm.

We consider a \textit{pool-based active learning} setting, in which the labeled dataset $\mathcal{L}$ is much smaller compared to a large pool of unlabeled items $\mathcal{U}$. We can acquire the label of any unlabeled item by querying an oracle (e.g. a human annotator) at a uniform cost per item. Suppose we are given a query budget $K$, such that we are allowed to query labels of a maximum of $K$ unlabeled items. An optimal active learner selects the set of $K$ items which maximizes the expected performance gain of the classification model upon retraining it with their labels. Selection of $K$ items for querying is done in an iterative manner such that in each iteration an instance $q$ is queried and the model is retrained with its label. 

\subsection{Optimization Problem}
We define our objective as finding $q$ unlabeled items which maximizes the likelihood of labeled instances while minimizing the uncertainty of label predictions of the unlabeled instances $\mathcal{U} \setminus q$. For any $q \in \mathcal{U}$ we estimate this objective of the model after training it on $q$. Training on an item $(x_q, y_q)$ updates model parameters $\hat{\theta}$ to $\hat{\theta}^{+(x_q, y_q)}$ such that 
\begin{equation}
\label{eq_eer}
\hat{\theta}^{+(x_q, y_q)} = \argmin_{\theta} \mathit{l} (f_\theta (G), Y_\mathcal{L}\cup  y_q),
\end{equation} where $\mathit{l}$ is the loss function (e.g. cross-entropy). We can write our objective as an optimization problem: 
\begin{equation}
q^* = \argmin_q \mathcal{E}(f_{\hat{\theta}^{+(x_q, y_q)}}), 
\end{equation}
where $\mathcal{E}$ is a cost function defined as
\begin{equation}
\label{eq_objective}
\mathcal{E}(f_{\hat{\theta}^{+(x_q, y_q)}})  = \mathit{l}(f_{\hat{\theta}^{+(x_q, y_q)}} (G), Y_{\mathcal{L}}) +  \mathbf{H}([f_{\hat{\theta}^{+(x_q, y_q)}} (G)]_{\mathcal{U} \setminus q}), 
\end{equation} in which we minimize the loss over labeled instances combined with $\mathbf{H}([f_{\hat{\theta}^{+(x_q, y_q)}} (G)]_{\mathcal{U} \setminus q})$, the entropy of unlabeled instances.

%As our objective we need to find query instances which maximizes the likelihood of  labeled instances while minimizing the entropy of missing labels.
%\begin{equation}
%\label{eq_objective}
%q^* = \argmin_q \mathcal{E}(f_{\hat{\theta}^{+(x_q, y_q)}} (G)), ~q \in V_\mathcal{U}.
%\end{equation}
%Our objective is to find instances $(x_q, y_q)$ which will minimize the error, quantified by the loss of the model on the rest of the unlabeled items $V_{\mathcal{U} \setminus q}$. If we can assume that the label $y_q$ can be known at the time of selecting it, then this objective can be expressed as
%\begin{equation}
%\label{eq_objective}
%q^* = \argmin_q \mathcal{E}(f_{\hat{\theta}^{+(x_q, y_q)}} (G), Y_{\mathcal{U} \setminus q}), ~q \in V_\mathcal{U}.
%\end{equation}

Since the label $y_q$ of an unlabeled instance $q$ is unknown, we compute the expected loss over all possible labels. We rewrite Equation \eqref{eq_objective} as
\begin{equation}
\label{eq_exp_objective}
\argmin_q \sum_{k=1}^C P(\hat{y}_q = k|G, Y_\mathcal{U} )) \mathcal{E}(f_{\hat{\theta}^{+(x_q, y_q=k)}}).
\end{equation}
In this case, we select the instance $x_q$ which minimizes the expected value of $\mathcal{E}$. $\hat{\theta}^{+(x_q, y_q=k)}$ denotes the parameters of a model trained with instance $q$ having the label $k$.

\subsection{Meta-learning Approach}
Since the label of an item $q \in \mathcal{U}$ is unknown, we use the posterior class probabilities $\hat{y}_q$ as a proxy for $y_q$. This approach requires training a separate model for each possible label of each unlabeled item ($N_\mathcal{U} \times C$). Training this many models is prohibitively time consuming. 
%and not practical for datasets containing a large number of items and classes.

To remedy this issue, we estimate the impact of a query $q$ with label $k$ ($y_q = k$) by training a model with label $\hat{y}_{q, k}$ upweighted by a small perturbation $\delta$ such that ($x_q, y_q = \hat{y}_q + \hat{y}_q \cdot \delta_{q,k}$), where $\delta_{q,k} \in \mathbb{R}$ is the perturbation added to label $k$. This idea is motivated by the use of perturbations in the feature space for finding training instances responsible for a given prediction \citep{koh2017influence}. In contrast, our objective is to find unlabeled instances which incur the greatest impact on the performance on test instances, once their labels are known. We re-purpose the use of perturbations to understand the impact an unlabeled instance $q$ may have on the model performance if it has the label $y_q$.
We rewrite Equation \eqref{eq_eer} as
\begin{equation}
\label{eqn:perturb_theta}
\hat{\theta}^{+(x_q, y_q=k)}= \argmin_{\theta} \mathit{l} (f_\theta (G), Y_\mathcal{L} \cup \hat{Y}_q \odot (1 + \delta_{q,k})).
\end{equation}
We quantify the impact of retraining the model with $(x_q, y_q)$ added to the labeled set as the change in loss
\begin{equation}
\Delta \mathcal{E}_{q, k} = \mathcal{E}(f_{\hat{\theta}^{+(x_q,y_q=k)}} ) - \mathcal{E}(f_{\hat{\theta}}),
\end{equation}
and the expected change of loss for querying the item $q$ by 
\begin{equation}
\label{eq_delta_lq}
\Delta \mathcal{E}_q = \sum_{k=1}^C P(\hat{y}_q = k|G, Y_\mathcal{L} )~\Delta \mathcal{E}_{q, k}.
\end{equation}
$P(y_q = k|G, Y_\mathcal{L} )$ is the posterior class probabilities of the current model $f_{\hat{\theta}}$ and it is estimated with 
\begin{equation}
	P(y_q = k|G, Y_\mathcal{L} ) = \text{softmax}(f_{\hat{\theta}}(G)).
\end{equation}
 When $\delta_{q,k}$ is arbitrarily small, this change can be computed as the gradient of the loss with respect to the label perturbation $\delta_{q,k}$, $\Delta \mathcal{E}_{q, k} \rightarrow \nabla_{\delta_{q,k}} \mathcal{E}(f_{\hat{\theta}_{q,k}}, Y_{\mathcal{U} \setminus q})$.
We rewrite Equation \eqref{eq_delta_lq} using gradient as
\begin{equation}
\label{eqn:expected_meta_grad}
\Delta \mathcal{E}_q  = \sum_{k=1}^C P(\hat{y}_q = k|G, Y_\mathcal{L} )~\nabla_{\delta_{q,k}} \mathcal{E}(f_{\hat{\theta}^{+(x_q,y_q=k)}}).
\end{equation}

The term $\Delta \mathcal{E}_q$ quantifies the impact of labeling a query $q$. This simplifies the active learning problem to finding the item corresponding to the minimum expected meta-gradient $\Delta \mathcal{E}_q$ (Equation \eqref{eqn:expected_meta_grad}) such that 
\begin{equation}
q^* = \argmin_q \Delta \mathcal{E}_q.
\label{eqn:grad_objective}
\end{equation}
Here, a negative valued expected meta-gradient corresponds to a model with lower expected loss. 
In other words, we need to find a query $q$ which maximizes the negative of the expected gradient ($-\Delta \mathcal{E}_q$).

Equation \eqref{eqn:perturb_theta} and Equation\eqref{eqn:grad_objective} form a bilevel optimization problem. Calculating the meta-gradients as in Equation \eqref{eqn:expected_meta_grad} involves a calculation of two gradients in a nested order, the inner one for optimizing the model parameters $\hat{\theta}_q$ for perturbed labels and the outer one for calculating the gradient with respect to the perturbation $\delta_{q,k}$. Therefore, the expected value of $\mathcal{E}$ indirectly depends on $\delta$ via $\hat{\theta}^{+(x_q, y_q=k)}$.  This is similar to the computation of meta-gradients in meta-learning approaches used for few-shot learning \citep{finn2017maml}. It should be noted that, unlike in few-shot learning, we calculate meta-gradients with respect to a perturbation added to the labels instead of differentiating with respect to model parameters. 
%Calculation of meta-gradient $\nabla_{\delta_{q,k}} \mathcal{E}(f_{\hat{\theta}_{q,k}}, Y_{\mathcal{U} \setminus q})$ involves calculating a gradient through a gradient.

Calculating $\Delta \mathcal{E}_q$ for each unlabeled node with Equation\eqref{eqn:expected_meta_grad} is inefficient for practical applications of this algorithm.We address this problem by selecting a subset of unlabeled items having higher prediction uncertainty to estimate the model uncertainty in Equation \eqref{eq_objective} and remaining unlabeled items as query items $\mathcal{Q}$. We add a small perturbation $\delta_{\mathcal{Q}} \in \mathbb{R}^{N_\mathcal{Q} \times C}$ to the labels of $\mathcal{Q}$ items and retrain the model with these perturbed labels. 
With vector notation we can rewrite Equation \eqref{eqn:perturb_theta} as
\begin{equation}
\label{eqn:vec_perturb_theta}
\hat{\theta}^{+(x_\mathcal{Q}, \hat{Y}_{\mathcal{Q}} \odot (1 + \delta))} = \argmin_{\theta} \mathit{l} (f_\theta (G), Y_\mathcal{L}\cup \hat{Y}_{\mathcal{Q}} \odot (1 + \delta)).
\end{equation}
Then we calculate the cost $\mathcal{E}$ and its gradient with respect to $\delta_\mathcal{Q}$ ($\nabla_{\delta_\mathcal{Q}})$. $\nabla_{\delta_\mathcal{Q}}$ is a real valued matrix, in which a row $q$ corresponds to an unlabeled instance $q \in \mathcal{Q}$  and a column $k$ corresponds to a label $k \in {1, \dots, C}$. For example, the gradient vector of query instance $q$ belonging to class $k$ can be expressed as $\nabla_{\delta_{q,k}} = [\nabla_{\delta_\mathcal{Q}}]_{[q, k]}$. We use the notation $[\nabla_{\delta_\mathcal{Q}}]_{[q, k]}$ to denote the element at $q$\textsuperscript{th} row and $k$\textsuperscript{th} column.

In our experiments, we use the top 10\% unlabeled items with the largest prediction entropy to estimate the model entropy and the rest of unlabeled items as $\mathcal{Q}$. 
%\kau{We do ablation studies with different values} 
Our algorithm is shown in Algorithm \ref{alg:meta_active}. We select the node corresponding to the minimal meta-gradient and retrieve its label from the oracle. We add this node and its label to the labeled set and retrain the model.

%\begin{figure}[htbp]
%	\centering
%	\subfigure[Amazon Computers]{\includegraphics[width=0.45\textwidth]{figures/amazon_electronics_computers_4_macro-f1}}
%	\hspace{5mm}
%	\subfigure[Amazon Photo]{\includegraphics[width=0.45\textwidth]{figures/amazon_electronics_photo_4_macro-f1}}
%	\caption{Macro-F1 score (test) of active learning algorithms with number of acquisitions. All datasets are queried with batches of 4 nodes.}
%	\label{fig:amazon_batch_4_macro-f1}
%\end{figure}

\subsection{The Importance of Exploration}
After each acquisition step, the classifier is trained on a limited number of labeled instances, which in turn are selected by the active learner. Hence, the selected labeled instances tend to bias towards instances evaluated to be  `informative' by the active learner. In MetAL, the active learner selects the instance which minimizes the meta-gradient. Therefore, the distribution of labeled instances is far from the true underlying distribution. The active learner cannot observe the consequences of selecting an instance which has lower `informativeness'. Therefore, it is desirable to query a few instances in addition to the ones maximizing our selection criteria. This step is known as `exploration' while selecting the instance maximizing the criteria is `exploitation'. Intuitively, an active learner should perform more exploration initially, so it can have a better view of the true distribution of data.

\begin{figure}
	\begin{center}
	\includegraphics[width=0.7\textwidth]{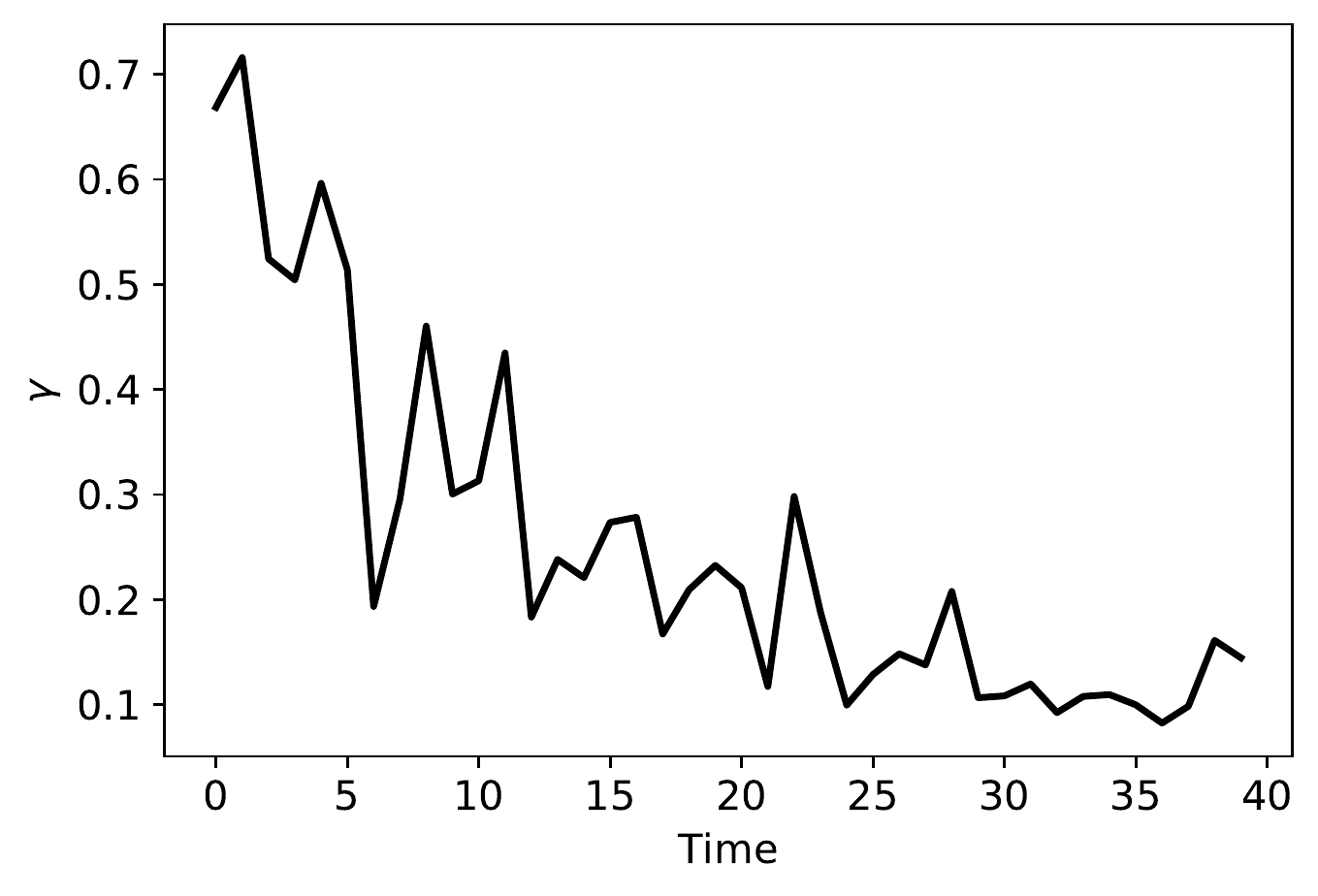}
	\caption{Variation of $\gamma_t$ over time. Average of 10 random samples of Beta($\alpha$, $\beta_t$) distribution. We set the value of $\alpha$ to 1 and increase $\beta_t$ with time $t$. }
	\label{fig:gamma_dist}
\end{center}
\end{figure}

This problem is known as \textit{exploration vs exploitation tradeoff} in sequential decision-making problems. Solving this tradeoff requires the learner to acquire potentially sub-optimal instances (i.e., exploration) in addition to the optimal ones. This problem is studied under the framework of multi-armed bandits problem \citep{lattimore2020bandit} (MAB). A multitude of approaches is used in solving online learning problems modeled as MAB problems. $\epsilon$-greedy, upper confidence bounds (UCB) \citep{auer2002ucb}, and Thompson sampling \citep{thompson1933likelihood} are few of the frequently used techniques. Influenced by count-based approaches proposed for MAB problems, we introduce a simple exploration term in addition to the exploitation performed using the meta-gradients. We define the exploration term of an instance $i$ as the logarithm of the number of unlabeled neighboring nodes of $i$. This term encourages the learner to sample nodes from less labeled neighborhoods. Since this term and the gradient calculated in Equation \eqref{eqn:expected_meta_grad} are on different scales, we normalize both of these quantities into $[0, 1]$ range and get $\phi_{exp}(i)$ and $\phi_{\nabla}(i)$ respectively. We linearly combine these normalized quantities to get the criterion for acquiring nodes as
\begin{equation}
\phi(i) = (1 - \gamma_t) \cdot \phi_{\nabla}(i) + \gamma_t \cdot \phi_{exp}(i),
\label{eqn:selection_criteria}
\end{equation}
where the exploration coefficient $\gamma_t$  is a hyper-parameter that balances exploration and exploitation. Setting $\gamma_t$ to 0 corresponds to pure exploration disregarding the feedback of the classifier (i.e. meta-gradient information). On the other hand, $\gamma_t = 1$ is equivalent to pure exploitation selecting the node with the minimum meta-gradient. We vary the value of $\gamma_t$ with time, such that more exploration is performed during the initial acquisition steps followed by more exploitation in later rounds. To achieve this effect, we assume $\gamma_t$ is sampled from a Beta distribution such that $\gamma_t \sim Beta(\alpha, \beta_t)$.  We linearly increase the value of $\beta_t$ over iterations of acquisitions to achieve the required effect of $\gamma_t$. As shown in \cite{zhang2017activedisc}, we observe smoother performance compared to setting the value of $\gamma_t$ deterministically. \autoref{fig:gamma_dist} shows how the value of $\gamma_t$ varies over time in average.

%Thompson sampling is a classic algorithm used in such problems to introduce exploration. It is shown that applying dropout \citep{srivastava2014dropout} at evaluation time can be interpreted as an implicit form of Thompson sampling \citep{riquelme2018deep}. We apply dropout with probability 0.5 when calculating the predictions of the model $f_{\hat{\theta}^{+(x_q, y_q)}}$ in Equation \eqref{eq_objective}. We perform ablations studies with different dropout rates to understand the impact of implicit exploration introduced by dropout in \autoref{fig:batch4_dropout}.

\begin{algorithm}
	\caption{MetAL: Meta Learning Active Node Classification.}
	\label{alg:meta_active}
	\begin{algorithmic}
		\STATE {\bfseries Input:} {Graph $G = (A, X)$,
			Query budget $K$, 
			Initial labels $Y_\mathcal{L}$}
		
		\STATE {\bfseries Output:} {An improved model}
		\STATE $\theta \leftarrow \text{train model on G with known labels } Y_\mathcal{L}$ 
		
		\FOR {$i \leftarrow 1$ to $n_q = K$}
		
		\STATE Calculate posterior class probabilities $\hat{Y}$ with the current model
		
		%		\FOR {$j \leftarrow 1$ to $p$}
		
		\STATE Sample a set of $N_\mathcal{Q}$ instances $\mathcal{Q}$ from $\mathcal{U}$
		%		\STATE Randomly initialize $\delta_{\mathcal{Q}_j}$
		
		\STATE Train a model with perturbed labels of $\mathcal{Q}$ instances with Equation \eqref{eqn:vec_perturb_theta}
		
		\STATE Calculate meta-gradient $\nabla_{\delta_{\mathcal{Q}}}$
		
		%		\ENDFOR
		
		\STATE Select the best instance $q^*$ using  Equation \eqref{eqn:selection_criteria}
		
		\STATE Query the oracle and retrieve the label $Y_{q^*}$
		\STATE Update label set $Y_\mathcal{L} \gets Y_\mathcal{L} \cup Y_{q^*}$
		
		\STATE Retrain the model $\theta \gets \argmin_{\theta} \mathit{l} (f_\theta (G), Y_\mathcal{L} )$
		
		\ENDFOR
		\STATE \bf Return $\theta$
	\end{algorithmic}
\end{algorithm}

\section{Experiments}
\subsection{Data}
We evaluate our proposed approach on 6 datasets belonging to different domains. CiteSeer, PubMed, and CORA~\citep{sen2008cora} are commonly used citation graphs. Each of these graphs is made of documents as nodes and citations as edges between them. If one document cites another, they are linked by an edge. Each node contains bag-of-words features of its text as its attributes. Co-author CS and Co-author Physics are co-authorship graphs constructed from Microsoft Academic Graph. Nodes are authors, two authors are linked by an edge if they have co-authored a paper. Node features correspond to the keywords of the papers authored by a particular author. An author's most active field of study is used as the node label. Amazon Computers is a subgraph of the Amazon co-purchase graph \citep{mcauley2015amazon}. Products are represented as nodes, two nodes are connected by an edge of those two products are frequently bought together. Node features correspond to product reviews encoded as bag-of-words. The product category is the node label.

For each dataset, we randomly select two nodes belonging to each label as the initial labeled set $V_\mathcal{L}$. We leave 5\%  of the rest of the unlabeled nodes as the test set. The remaining unlabeled nodes $V_\mathcal{U}$  qualify to be queried. The size of the initial labeled set and its size as a  fraction of the total nodes (labeling rate) are shown in \autoref{tbl:data}.

\begin{table}
	\centering
	\caption{Dataset statistics. Labeling rate as a percentage of total nodes is shown within brackets.}
	\label{tbl:data}
	\begin{tabular}{lrrrr}
		\hline
		Dataset  & Nodes & Classes & Features  & Labels (\%) \\
		\hline
		CiteSeer & 2110            & 6                 & 3703 & 12 (0.56)                   \\
		PubMed   & 19717           & 3                &  500 & 6 (0.03)       \\
		CORA   &  2485            & 7               &  1433 & 14 (0.56) \\
		 Amazon Computers   &  13752            & 10            &  767  & 20 (0.14) \\
%		 Amazon Photo   &  7650            & 8                & 16 (0.21) \\
		 Co-author Physics  &  34493  & 5  &   8415  & 10 (0.03) \\
		 Co-author CS  &  18333           & 15             &  6805 & 30 (0.16) \\
		\hline           
	\end{tabular}
\end{table}

\subsection{Model}
We evaluate the effectiveness of MetAL, the proposed algorithm using a two-layer
GCN model~\citep{kipf2017gcn} with 64 hidden units and SGC \citep{sgc2019}, a simplified GNN architecture that does not include a hidden layer and nonlinear activation functions. In all experiments, we use the default hyper-parameters used in GNN literature (e.g. learning rate = 0.01). We do not perform any dataset-specific hyper-parameter tuning since hyper-parameter tuning while training a model with AL can lead to label inefficiency~\citep{ash2019badge}. We use the following algorithms in our comparison: 
\begin{itemize}
	\item \textbf{Random:} Selects an unlabeled node randomly.
	\item \textbf{PageRank (PR):} Selects the unlabeled node with the largest PageRank centrality value.
	\item \textbf{Degree}: Selects the unlabeled node with the largest degree centrality value.
	\item \textbf{Entropy:} Calculates the entropy of predictions of the current model over unlabeled nodes and select the node corresponding to the largest entropy value. 
	\item \textbf{AGE \citep{age2017cai}:} Selects the node which maximizes a linear combination of three metrics: PageRank centrality, model entropy and information density.
	\item \textbf{BALD \citep{gal2017bald, houlsby2011bald}:} Selects the node which has the the largest mutual information value between predictions and model posterior.
	\item \textbf{MetAL:} This is our proposed algorithm. We select the node maximizing the quantity in Equation \eqref{eqn:selection_criteria}
\end{itemize}

Here, entropy and BALD are uncertainty-based acquisition functions. For computing entropy, mutual information in BALD, and posterior class probabilities predicted by the current model $P(\hat{y}_q=k | G, Y_{\mathcal{L}})$ in MetAL, we use 20 iterations of MC-dropout to approximate a Bayesian model~\citep{gal2016dropout}. In contrast, centrality metrics such as PageRank and degree centrality can be considered as heuristics for selecting `influential' instances in a graph dataset. The sequence of acquisitions is determined only based on the graph structure and does not depend on the features of instances nor the current set of labeled instances. 

We acquire the label of an unlabeled node and retrain the GNN model by performing 50 steps of adam optimizer~\citep{kingma2014adam}. We perform 40 acquisition steps and repeat this process on 10 different randomly initialized training and test splits for each dataset. We report the average F1 score (Macro-averaged) over the test sets in each experiment. In most cases, average accuracy follows a similar trend. In MetAL, we execute 10 steps of gradient descent with momentum as the inner optimization loop and then we calculate the meta-gradient matrix. 

\begin{figure}[htbp]
	\centering
	\subfigure[CiteSeer]{\includegraphics[width=0.48\textwidth]{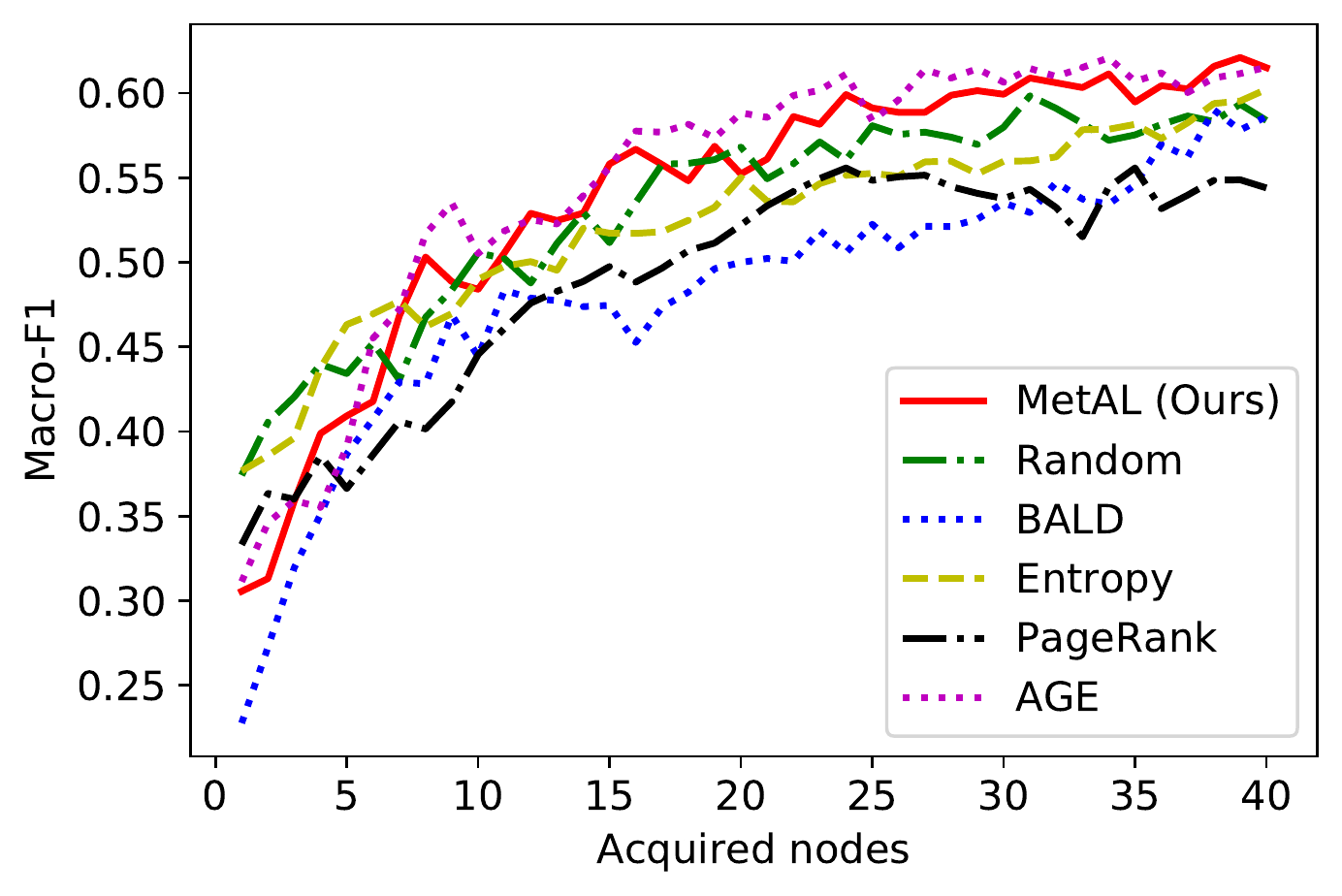}}
%	\hspace{5mm}
	\subfigure[PubMed]{\includegraphics[width=0.48\textwidth]{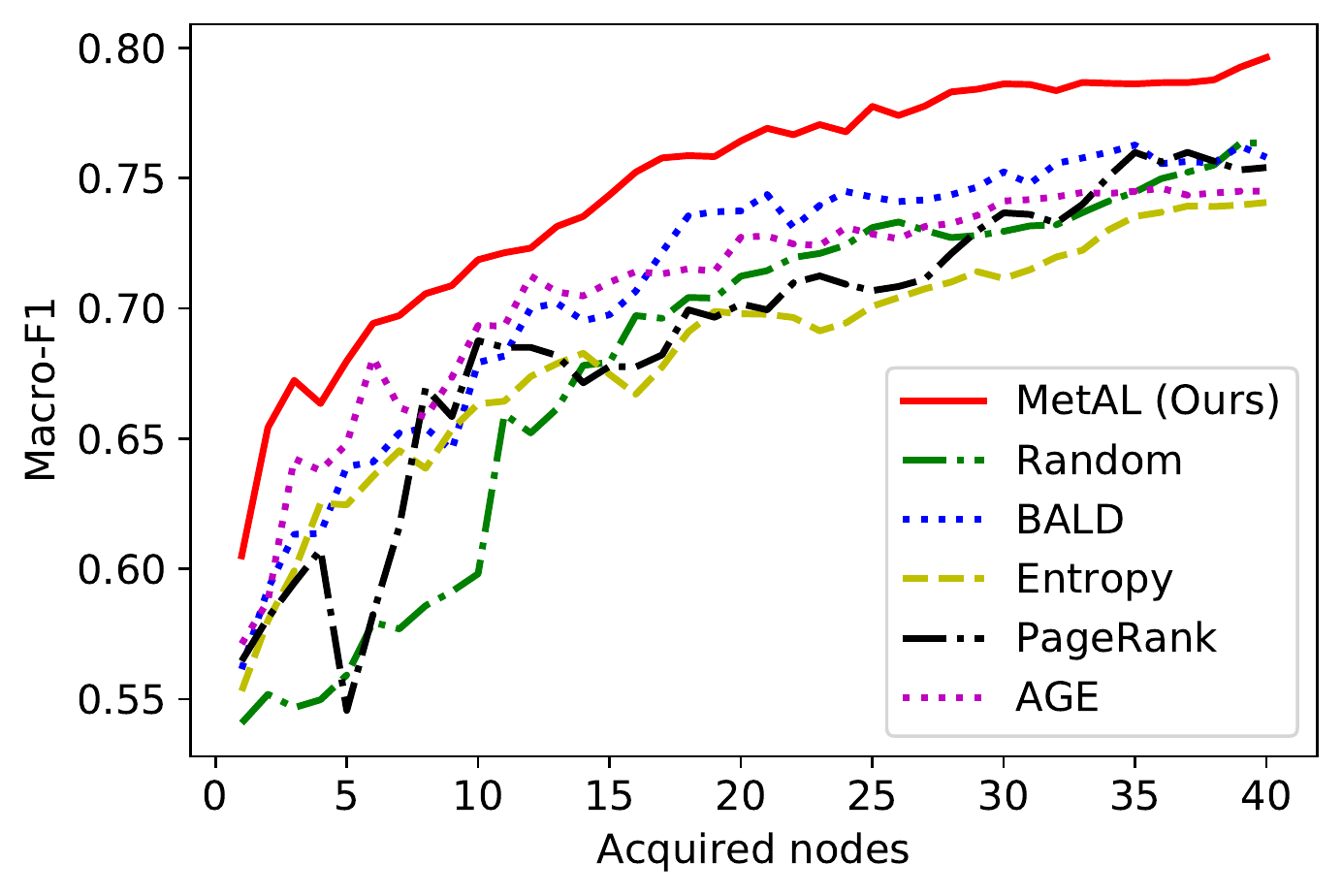}}
%	\hspace{5mm}
	\subfigure[CORA]{\includegraphics[width=0.48\textwidth]{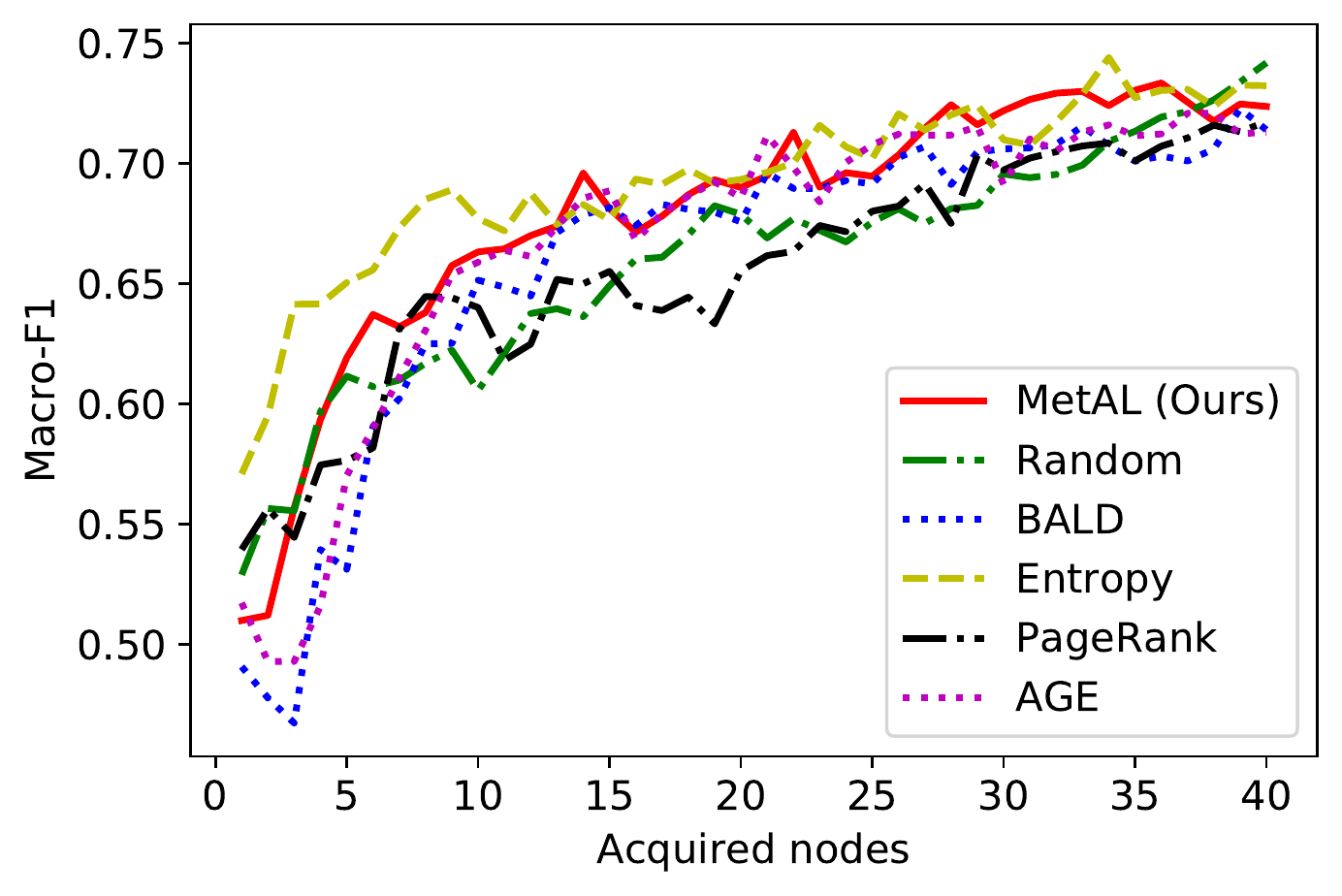}}
%	\hspace{5mm}
	\subfigure[Amazon Computers]{\includegraphics[width=0.48\textwidth]{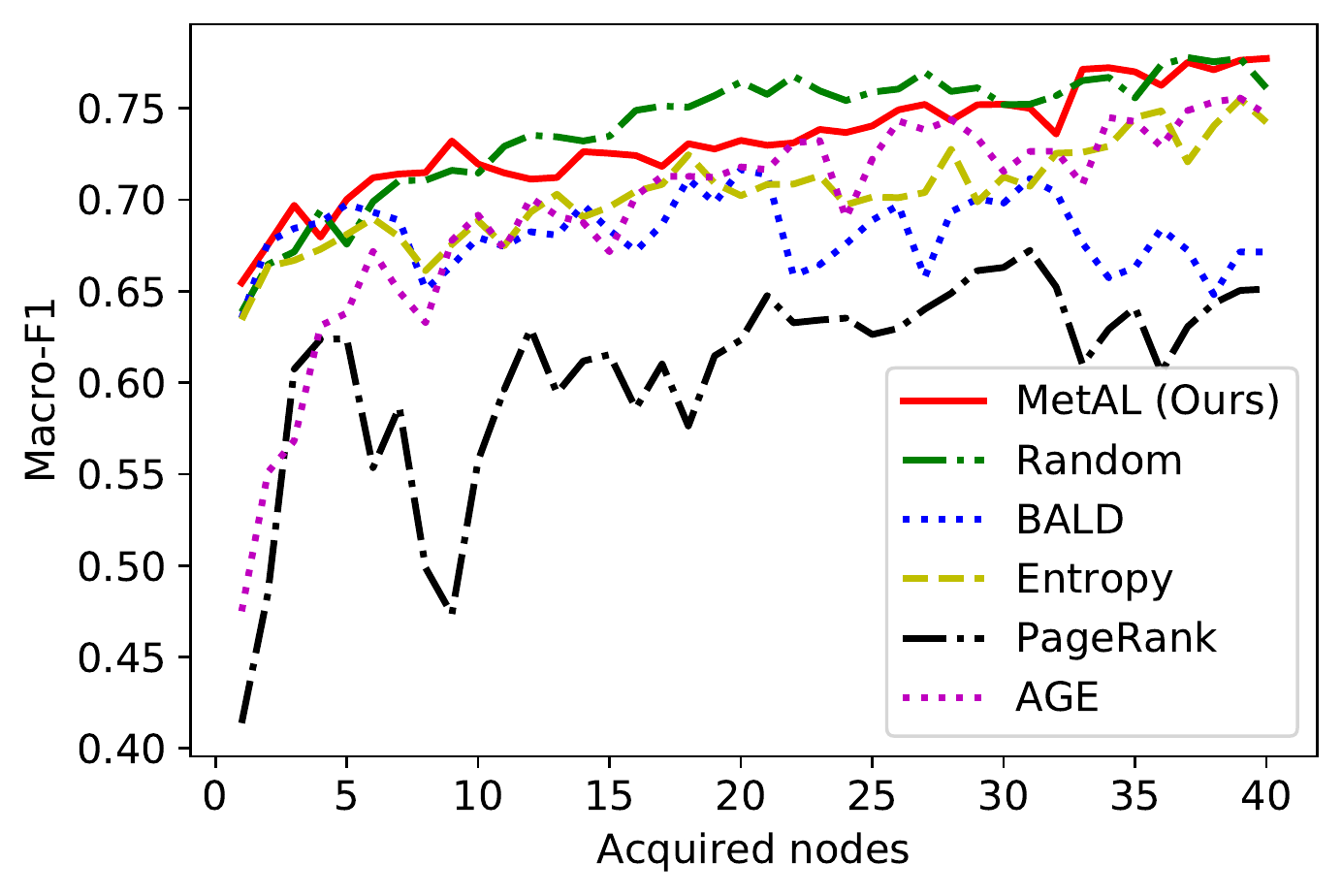}}
	\subfigure[Co-author Physics]{\includegraphics[width=0.48\textwidth]{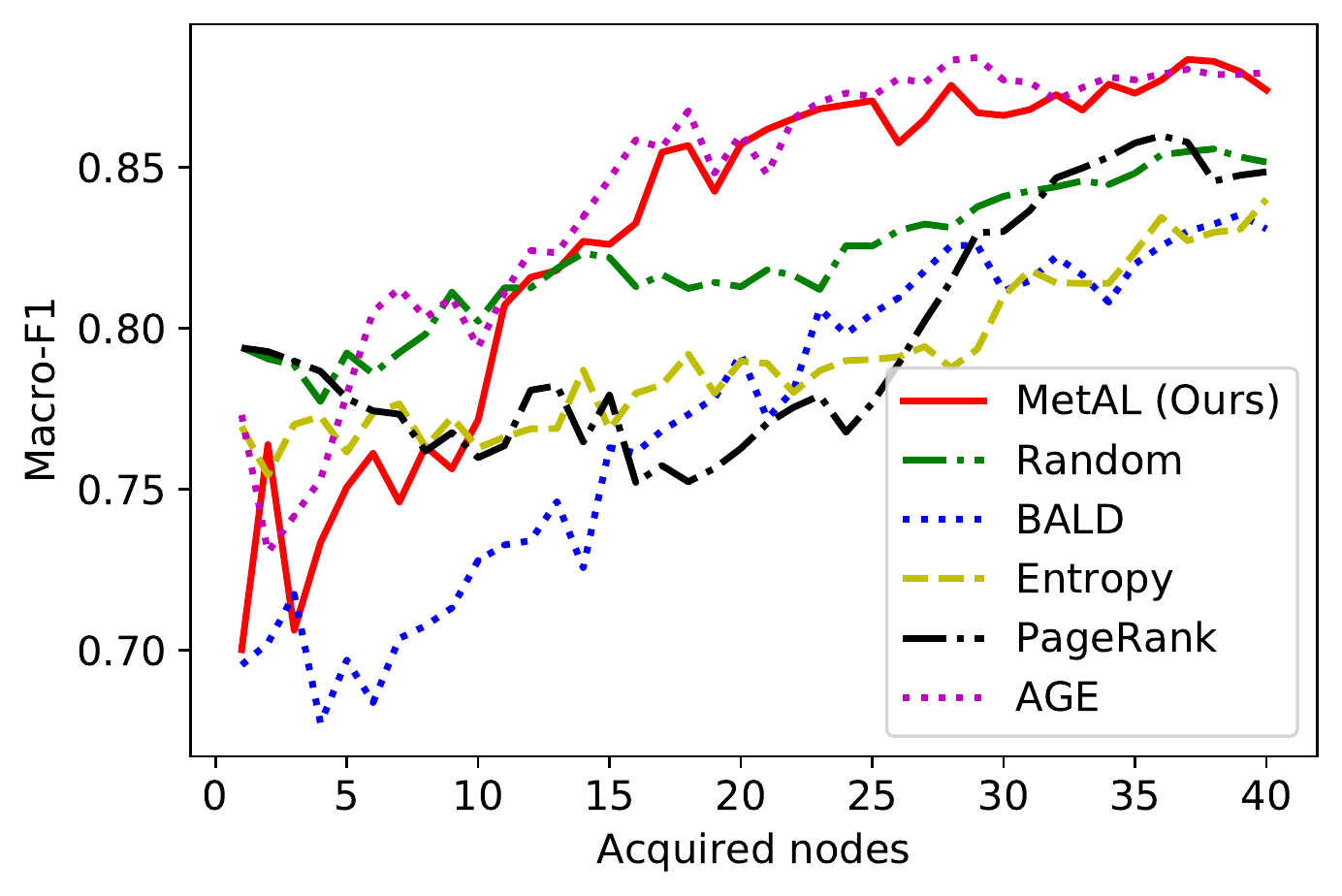}}
		\subfigure[Co-author CS]{\includegraphics[width=0.48\textwidth]{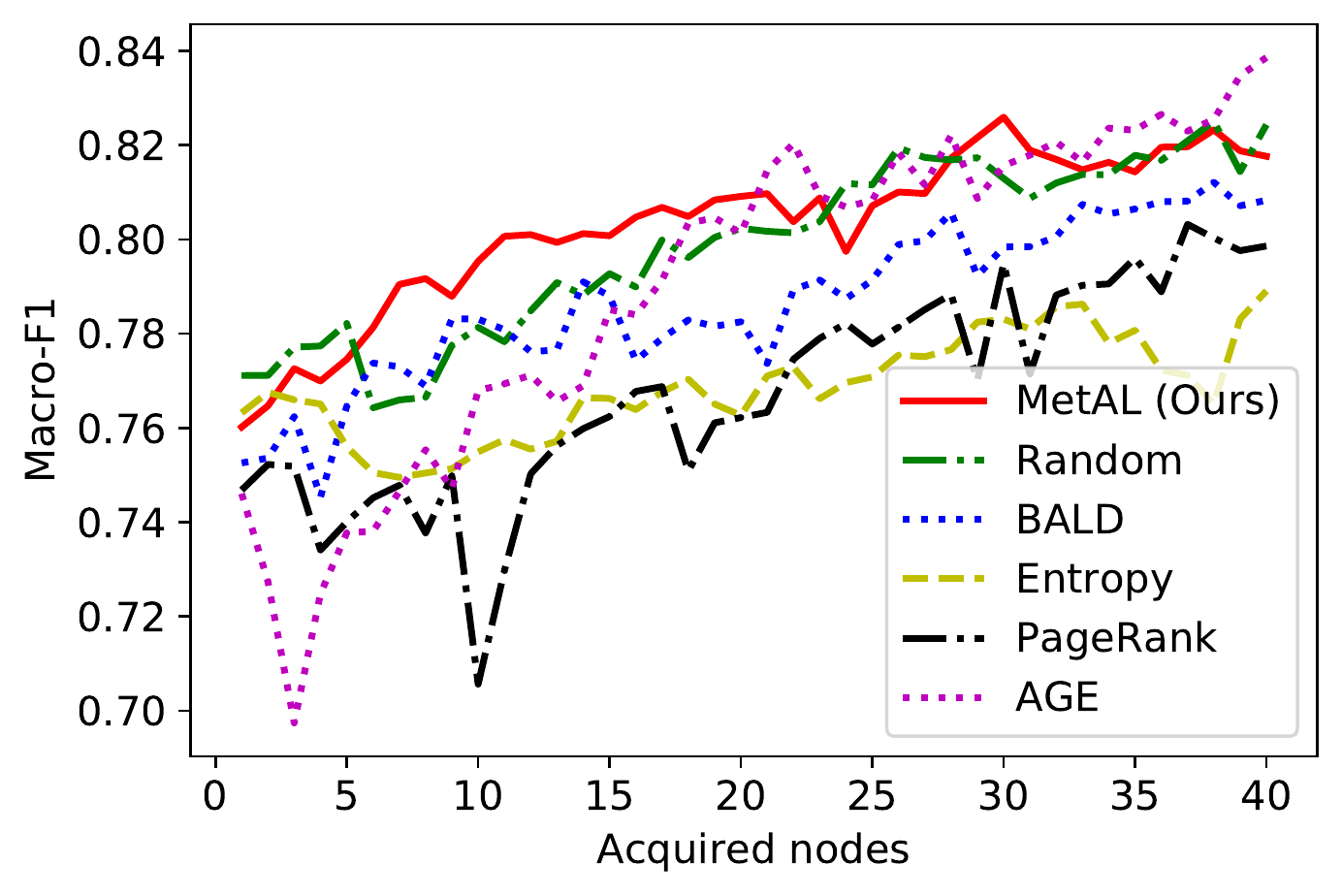}}
%	\hspace{5mm}
%	\hspace{5mm}
%	\subfigure[Amazon Photo]{\includegraphics[width=0.48\textwidth]{figures/amazon_electronics_photo_1_gcn.pdf}}
	\caption{Performance of active learners with a 2-layer GCN model as the node classifier. Macro-F1 score (test) of active learning algorithms with number of acquisitions. }
	\label{fig:batch1_gcn}
\end{figure}

\begin{figure}[htbp]
	\centering
	\subfigure[CiteSeer]{\includegraphics[width=0.48\textwidth]{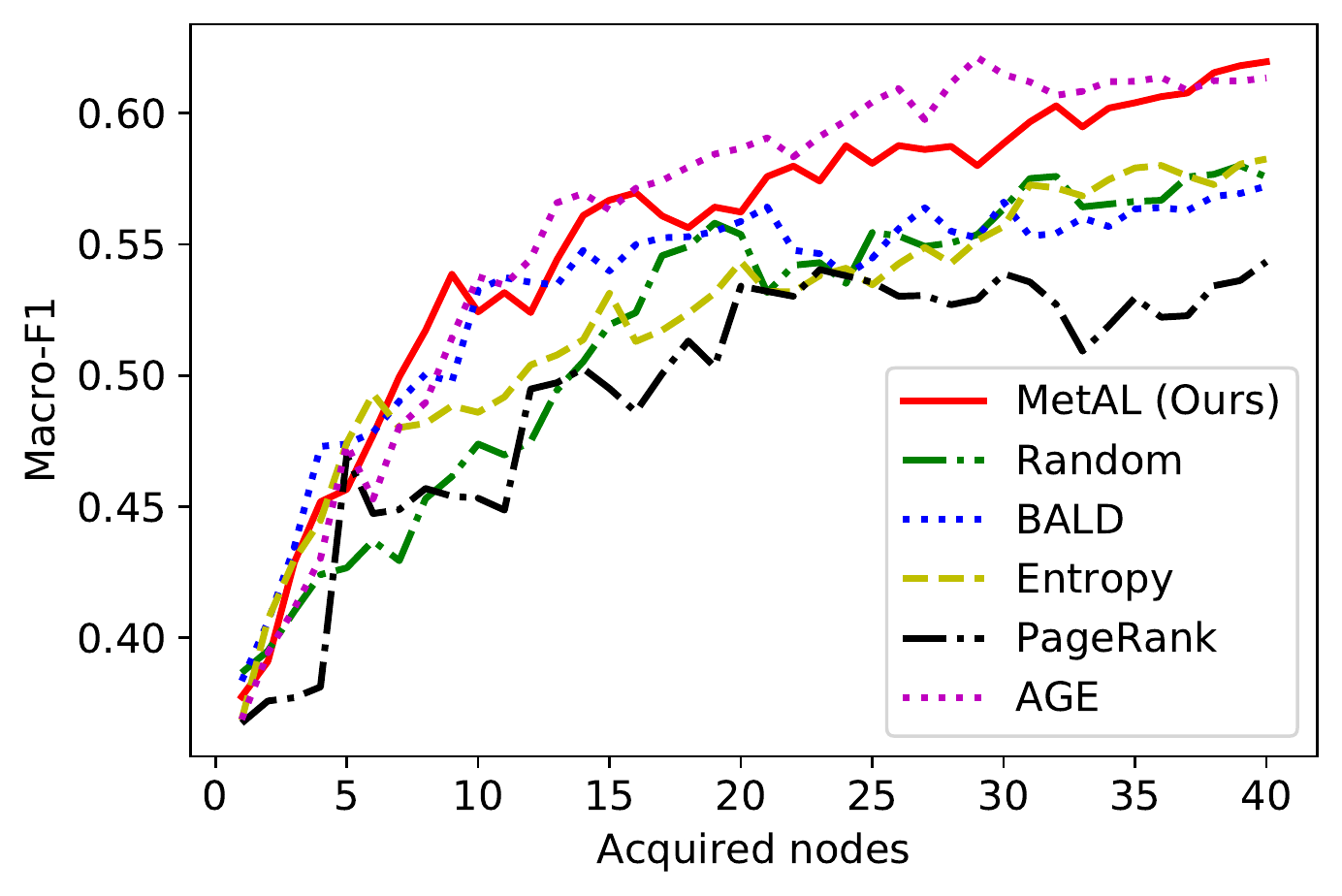}}
	%	\hspace{5mm}
	\subfigure[PubMed]{\includegraphics[width=0.48\textwidth]{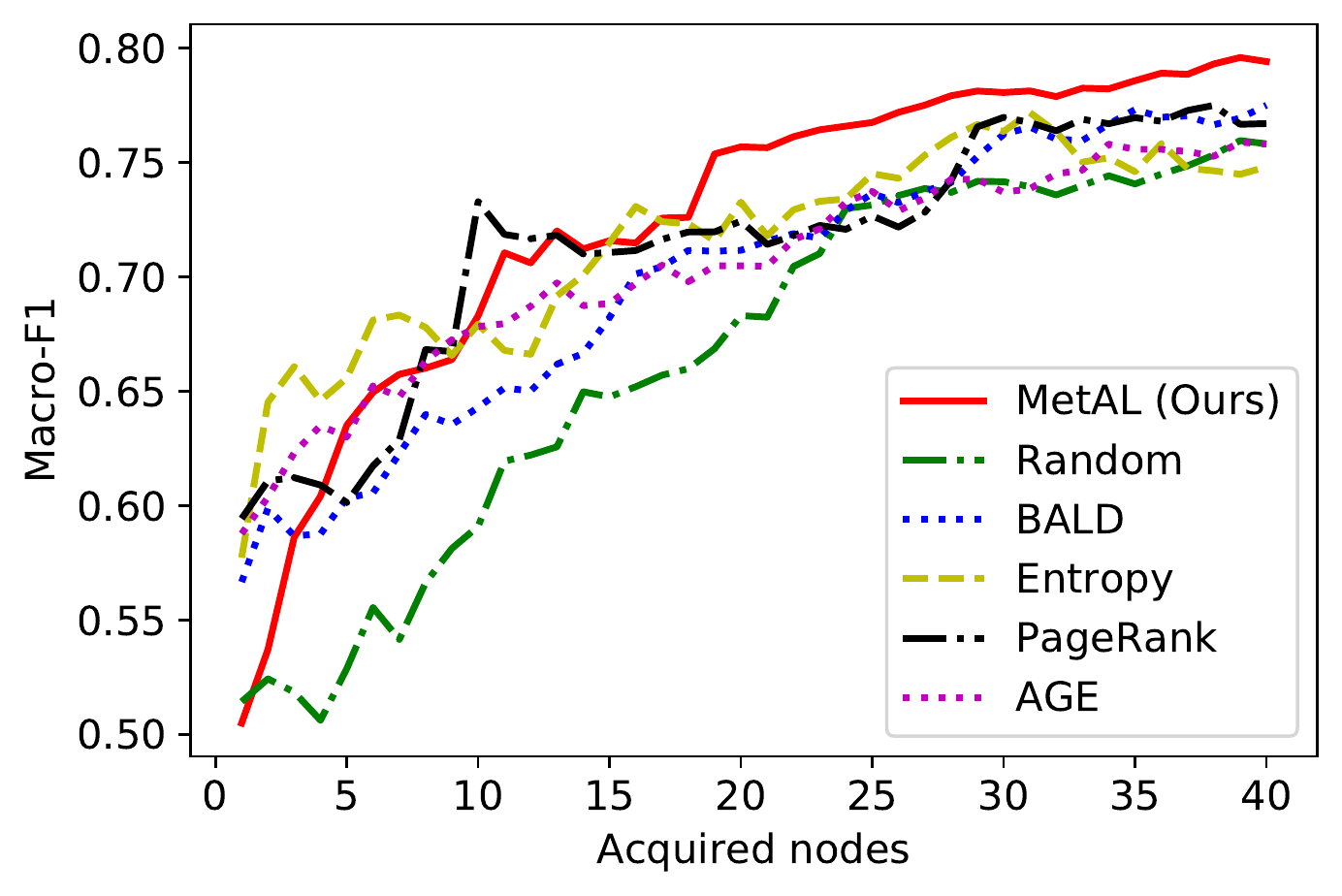}}
	%	\hspace{5mm}
	\subfigure[CORA]{\includegraphics[width=0.48\textwidth]{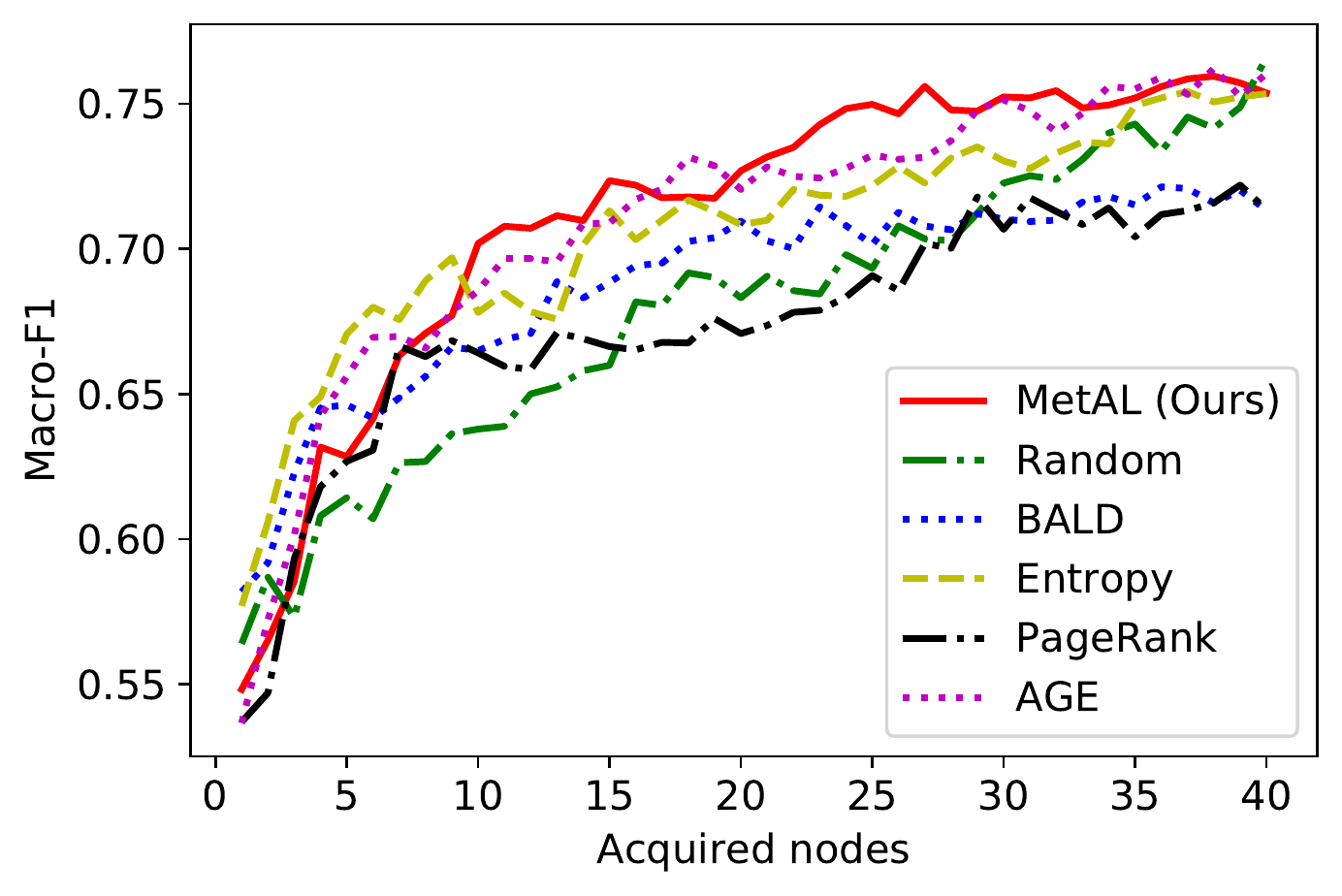}}
	%	\hspace{5mm}
	\subfigure[Amazon Computers]{\includegraphics[width=0.48\textwidth]{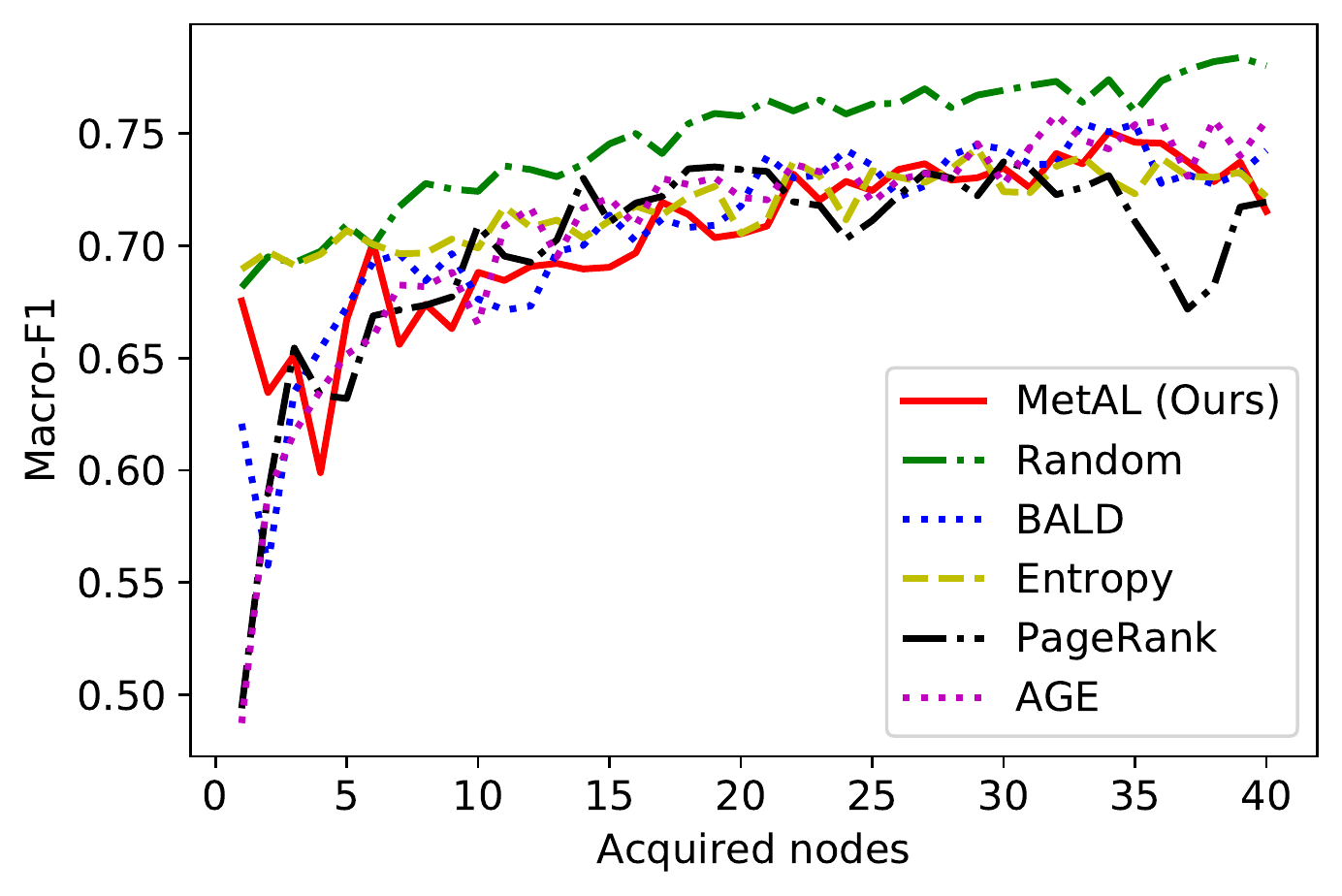}}
	\subfigure[Co-author Physics]{\includegraphics[width=0.48\textwidth]{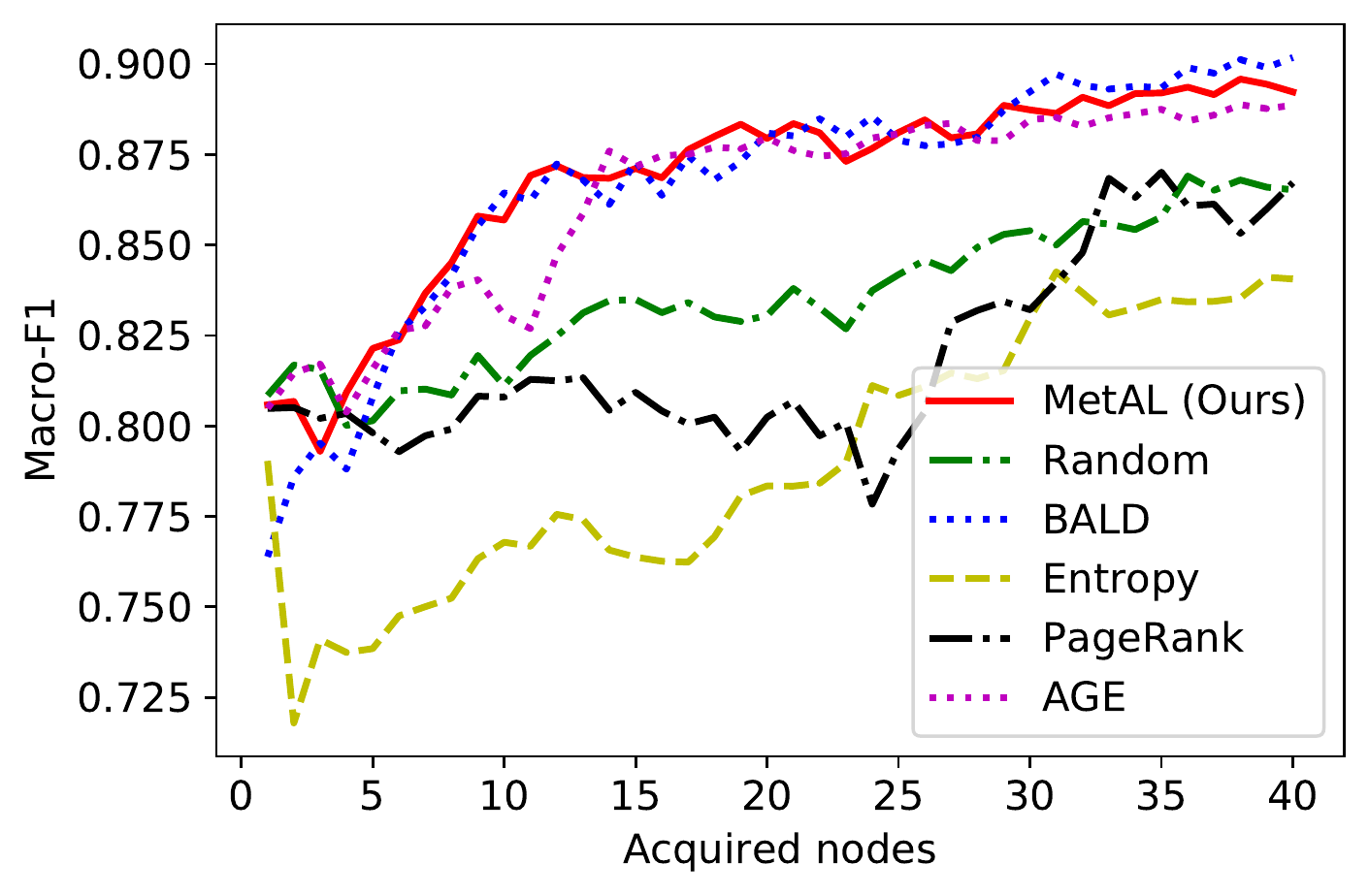}}
	\subfigure[Co-author CS]{\includegraphics[width=0.48\textwidth]{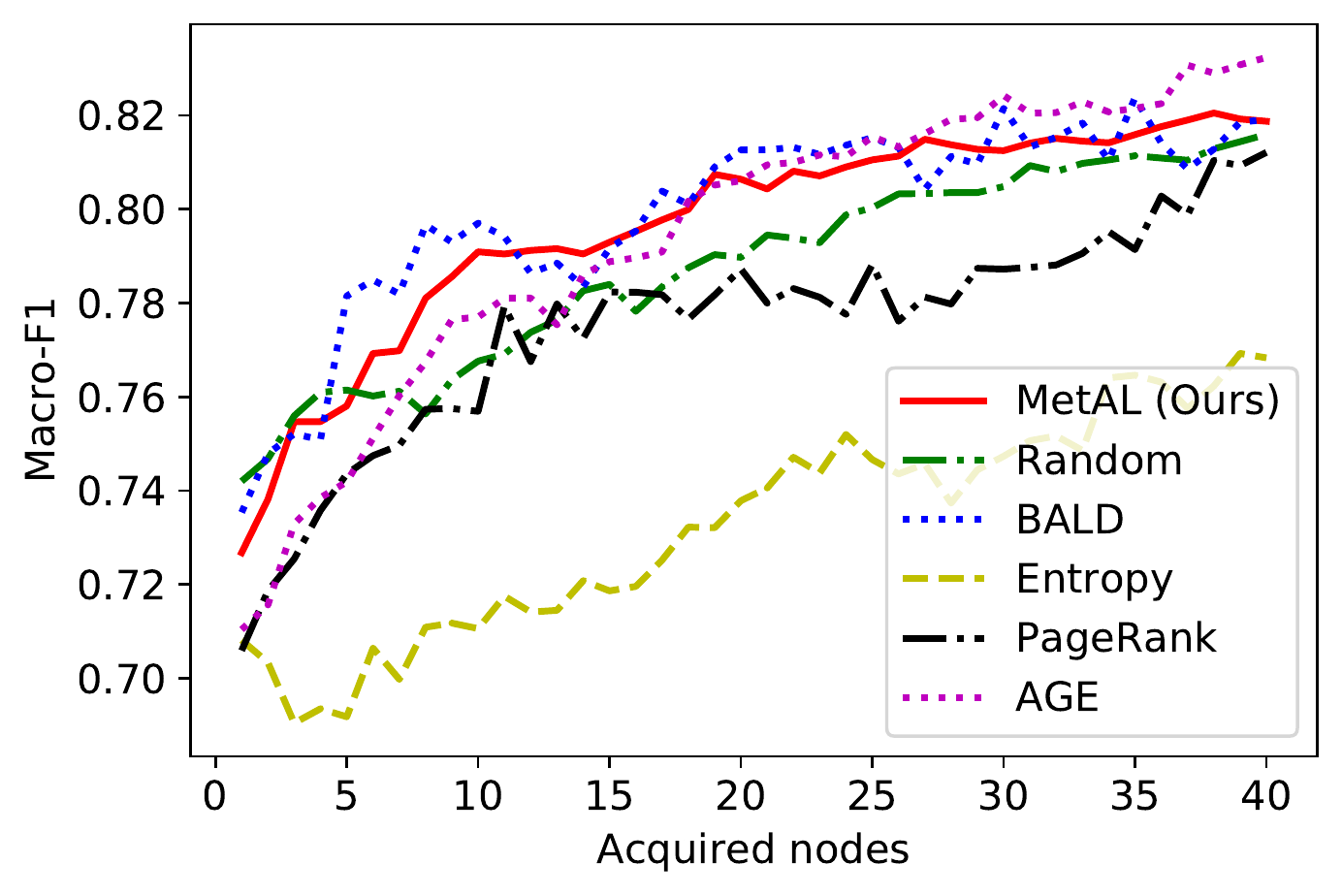}}
	%	\hspace{5mm}
	%	\hspace{5mm}
	%	\subfigure[Amazon Photo]{\includegraphics[width=0.48\textwidth]{figures/amazon_electronics_photo_1_gcn.pdf}}
	\caption{Performance of active learners with an SGC model as the node classifier. Macro-F1 score (test) of active learning algorithms with number of acquisitions. .}
	\label{fig:batch1_sgc}
\end{figure}

\begin{figure}[htbp]
	\centering
	\subfigure[CiteSeer]{\includegraphics[width=0.48\textwidth]{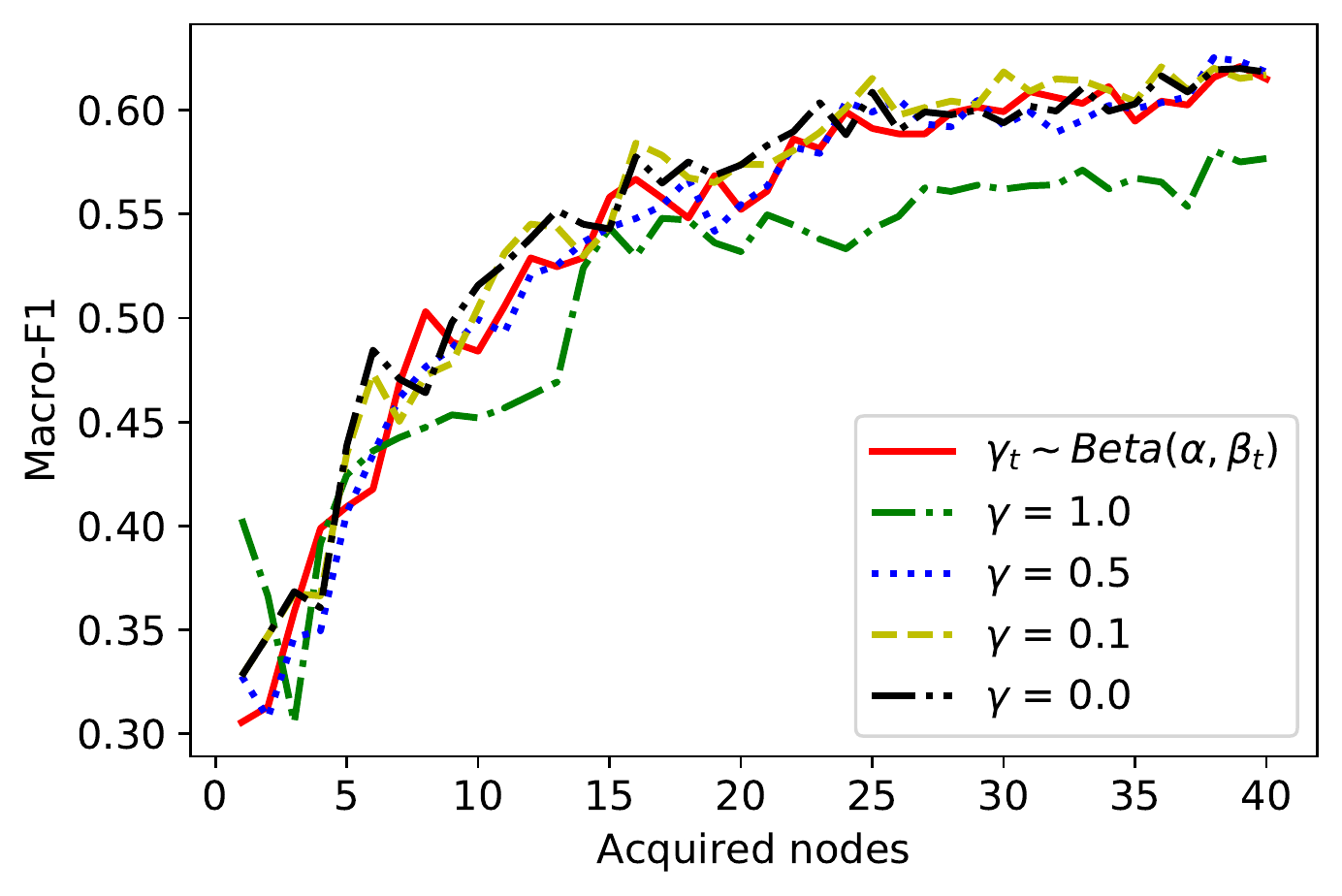}}
	%	\hspace{5mm}
	\subfigure[PubMed]{\includegraphics[width=0.48\textwidth]{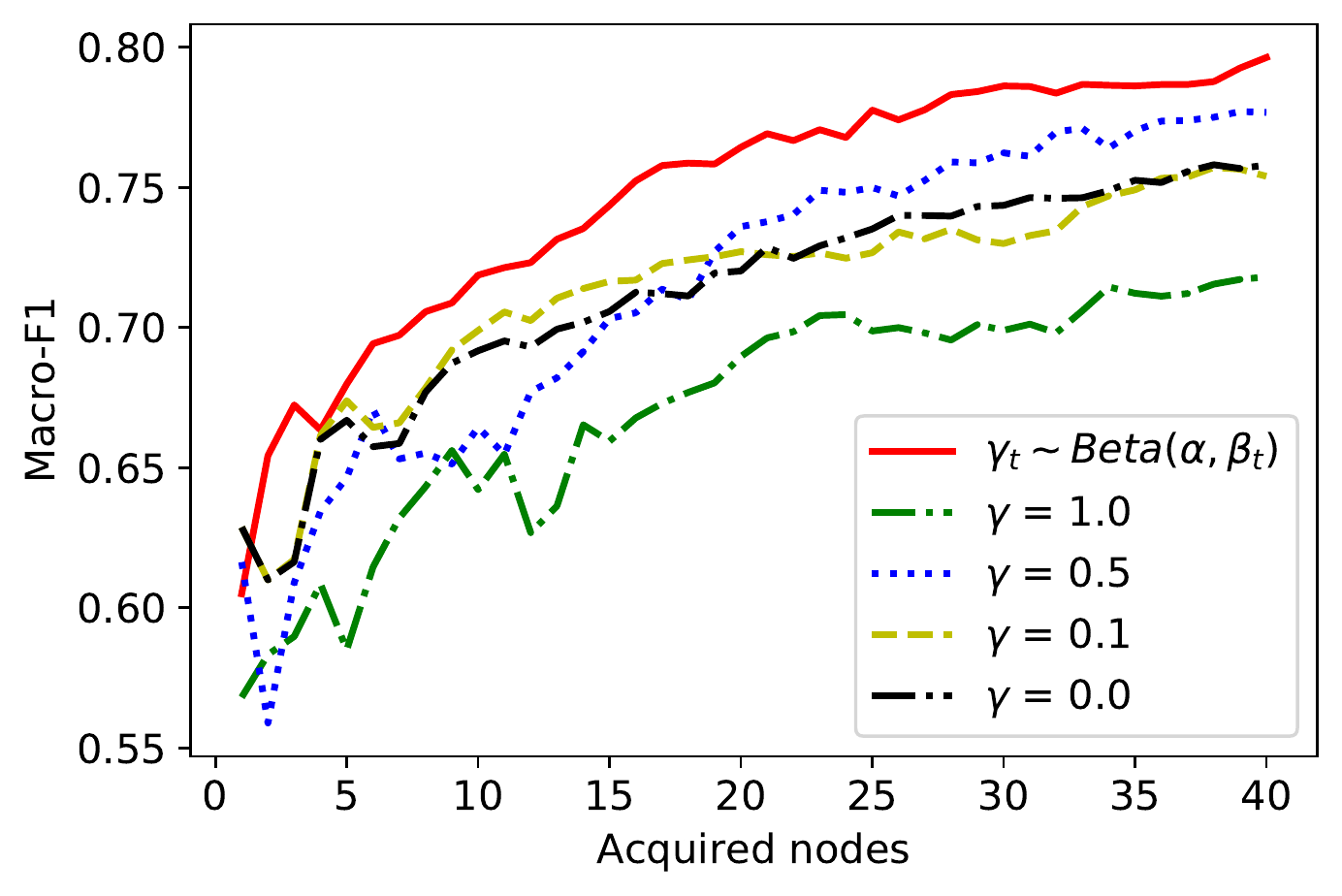}}
	%	\hspace{5mm}
	\subfigure[CORA]{\includegraphics[width=0.48\textwidth]{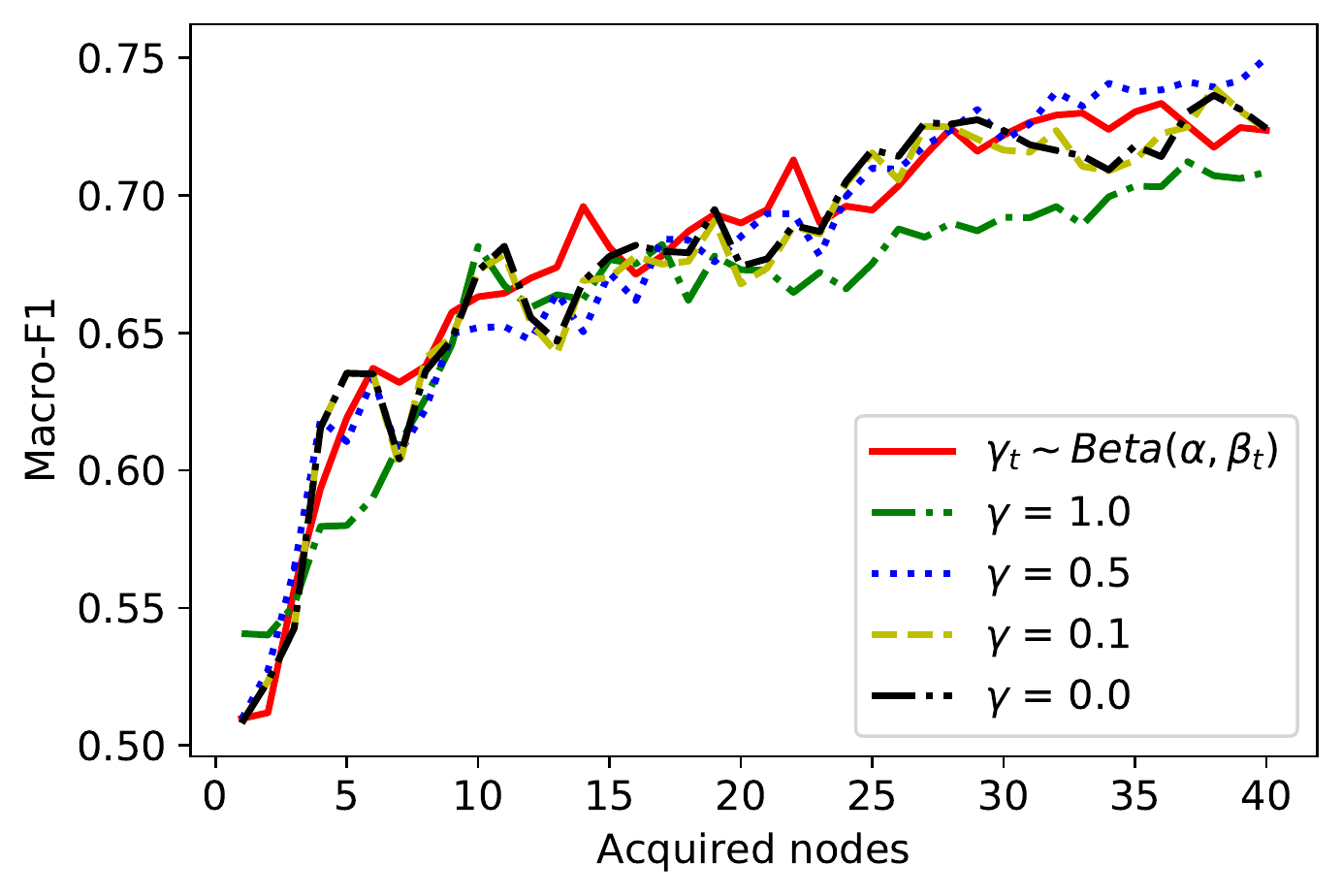}}
	\subfigure[Amazon Computers]{\includegraphics[width=0.48\textwidth]{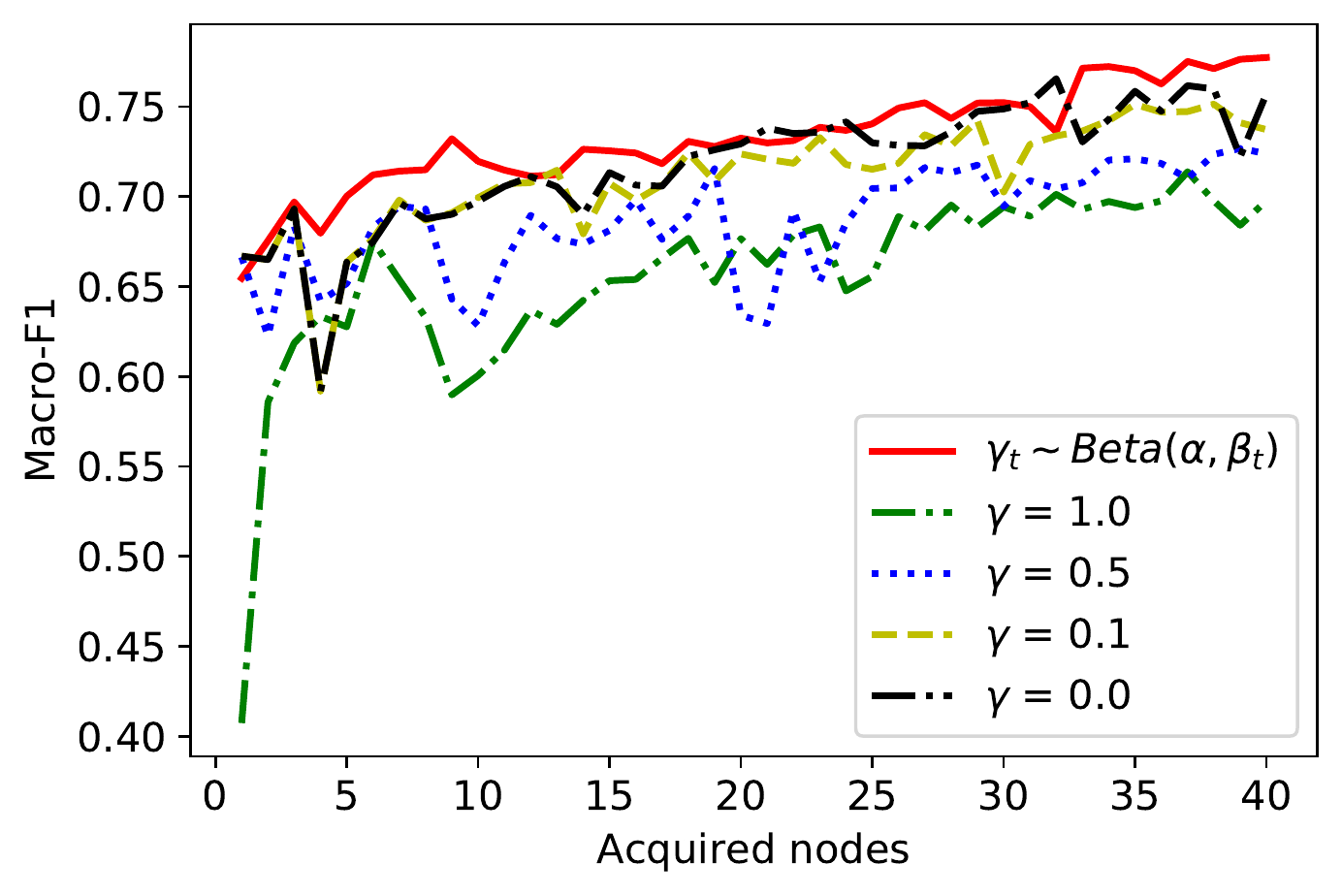}}
	\subfigure[Co-author CS]{\includegraphics[width=0.48\textwidth]{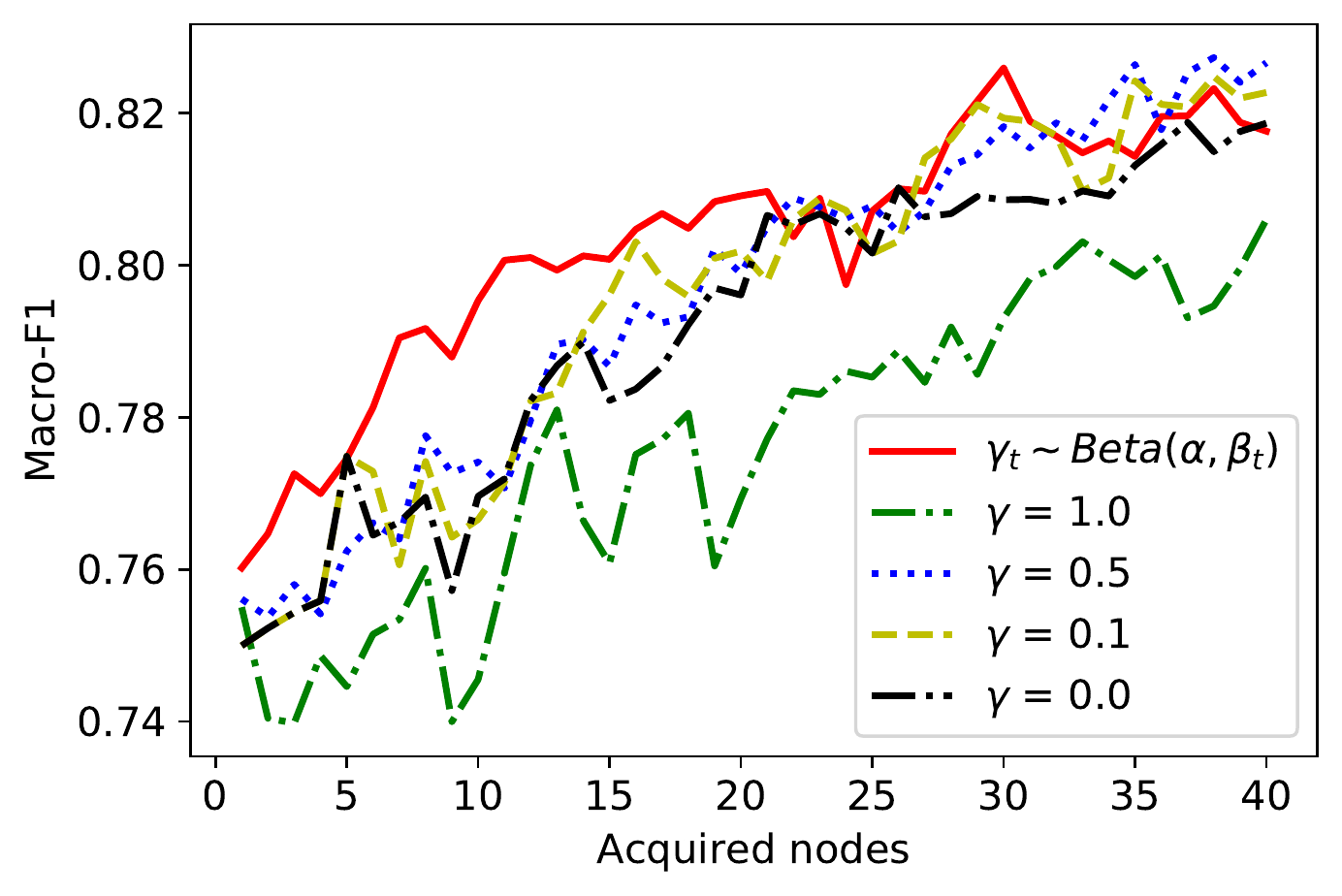}}
	\subfigure[Co-author Physics]{\includegraphics[width=0.48\textwidth]{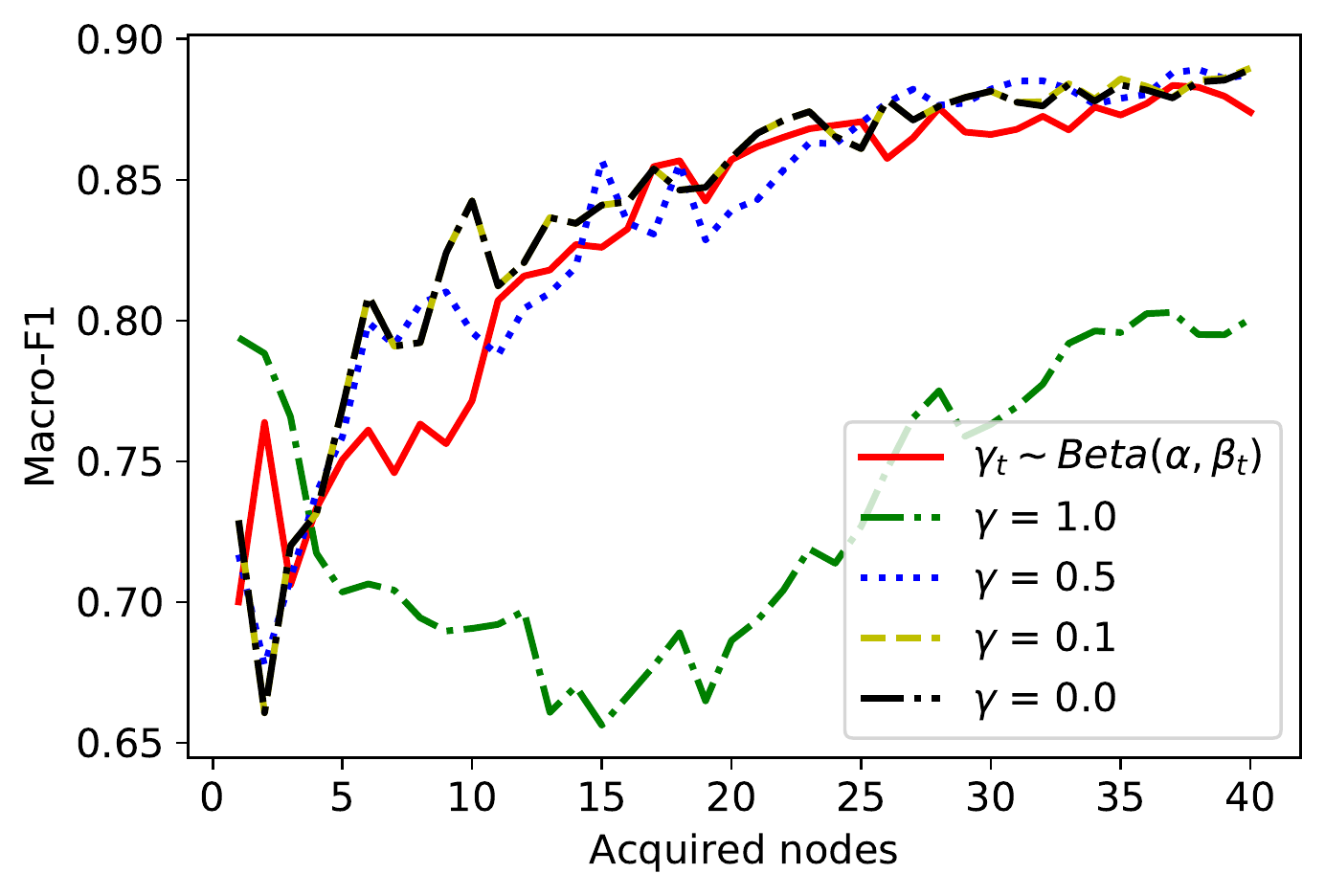}}
	%	\hspace{5mm}
	%	\subfigure[Amazon Photo]{\includegraphics[width=0.48\textwidth]{figures/amazon_electronics_photo_1_gcn.pdf}}
	\caption{The importance of  exploration coefficient $\gamma$. Our algorithm MetAL is run on these datasets with different values of $\gamma$: 0.1, 0.1, 0.5, and 1.0. The performance of MetAL with a fixed $\gamma$ value is compared against the performance when it is sampled from a Beta distribution which is time-dependent.}
	\label{fig:batch1_ablation}
\end{figure}

%\begin{figure}[htbp]
%	\centering
%	\subfigure[CiteSeer]{\includegraphics[width=0.3\textwidth]{figures/citeseer_4_dropout.pdf}}
%	\hspace{5mm}
%	\subfigure[PubMed]{\includegraphics[width=0.3\textwidth]{figures/pubmed_4_dropout.pdf}}
%	\hspace{5mm}
%	\subfigure[CORA]{\includegraphics[width=0.3\textwidth]{figures/cora_4_dropout.pdf}}
%	%	\label{fig:batch_4_macro-f1}
%	\caption{Ablation study to understand the importance of dropout rate for exploration in batch acquisitions. Macro-F1 score (test) of active learning algorithms with number of acquisitions. A two-layer GCN is used as the node classifier. A batch of 4 instances are acquired at once.}
%	\label{fig:batch4_dropout}
%\end{figure}

\section{Results and Discussion}
\label{sec:results}

\subsection{Comparison of AL Strategies}

%\kau{AGE is good at solving the cold-start problem at the initial acquisition rounds.}
In \autoref{fig:batch1_gcn} we observe that MetAL contributes to the best performance when the GCN model is used as the node classifier. We do not show the performance for degree centrality sampling for the clarity of visualizations since it exhibits the worst performance compared to all other acquisition functions. \autoref{fig:batch1_sgc} shows that MetAL works similarly with SGC as the classifier as well. However, we observe that the performance of SGC on some datasets is inferior compared to the GCN model. Lack of a hidden layer and nonlinear activation functions can be the reason contributing to reduced performance. Even though PageRank centrality is proposed as a heuristic for acquiring nodes of a graph in previous work \cite{age2017cai}, we observe that its performance is inferior on larger graphs such as Amazon and co-authorship graphs. The performance of uncertainty-based active learners (entropy and BALD) is not consistent over different datasets. It is interesting that MetAL consistently outperforms AGE, the graph-specific AL benchmark without relying on time-consuming clustering algorithms. As one of its constituent criteria, AGE computes an information density measure using the learned features of the GNN model. This step depends on clustering the unlabeled instances and then calculating the euclidean distance to the cluster centers. This process is time consuming as evident in \autoref{tbl:run_time}.

\autoref{fig:batch1_ablation} shows the results of the ablation studies we perform to understand the impact of the exploration coefficient $\gamma_t$. Here, we run the acquisition step in Equation \eqref{eqn:selection_criteria} with different $\gamma_t$ values: 0, 0.1, 0.5, and 1.0. Time-dependent  $\gamma_t$ sampled from a Beta distribution works the best on most datasets. Notably, selecting nodes solely based on the meta-gradient values ($\gamma_t$ = 0) results in competitive results in most cases. However, pure exploration ($\gamma_t$ = 1) results in inferior performance. This demonstrates that our proposed meta-gradient criterion is successful in finding `informative' instances for labeling. However, this experiment shows that performance can be further improved by adaptively updating $\gamma_t$ based on the feedback of the active learner.

\subsection{Running Time}
\autoref{tbl:run_time} lists the execution time each algorithm spends to acquire a set of 40 unlabeled instances on average. Even though our proposed approach MetAL consumes additional time compared to uncertainty-based algorithms, it is several times faster than the graph-specific baseline AGE. For example, MetAL is 20 times faster than AGE for the co-author Physics dataset. Further, the ultimate goal of applying AL is to reduce total human time spent on labeling instances. MetAL achieves this key objective at the cost of slightly increased acquisition time.  

\begin{table}
	\centering
	\caption{Running time (seconds): average time taken to acquire 40 unlabeled instances. We run all experiments on a single Nvidia GTX 1080-Ti GPU.}
	\label{tbl:run_time}
	\begin{tabular}{llrrrrrr}
		\hline
		Classifier & Dataset  & Random & Entropy & PR & AGE & BALD & MetAL \\
		\hline
		GCN & CiteSeer  & 4.2 & 4.8 &  4.8 & 21.5 & 4.8 & 9.7 \\
		 & PubMed   & 6.9 & 7.6 & 25.4 & 1125.9 & 7.9 & 34.6    \\
		& CORA & 4.2 & 4.5 & 4.6 & 26.8 & 4.5 & 9.8  \\
		& Co-author CS & 20.4  & 22.3 & 40.8 & 2154 .2 & 23.7 & 61.3\\
		& Co-author Phy. &  46.1 & 50.5 & 116.4 & 2436.9 & 50.8& 125.4 \\
		& Amazon Comp. & 17.5 & 19.1 & 31.8 & 1688.9 & 19.2 & 45.2 \\
		\hline
		SGC & CiteSeer  & 1.7 & 1.9 & 2.1 & 18.3 & 1.9 & 5.4 \\
		& PubMed   & 2.0 & 2.2 & 20.0 & 1229.2 & 2.2 & 30.6    \\
		& CORA & 1.3 &  1.8 & 1.8 & 23.7 & 1.9 & 5.5  \\
		& Co-author CS &16.8 & 19.8 & 33.2 & 2098.2 & 19.8 & 48.6 \\
		& Co-author Phy. & 35.6 & 40.7 & 90.4 & 2232.3 & 40.8 & 97.0 \\
		& Amazon Comp. &  2.2 & 2.5& 17.2 & 1134.6 & 2.5 & 22.0 \\
		\hline           
	\end{tabular}
\end{table}

\section{Related Work}

\subsection{Graph  Neural Networks (GNNs)}
GNNs \citep{li2015gated, kipf2017gcn,  sgc2019} achieve state-of-the-art performance on the node classification problem providing a significant improvement over previously used embedding algorithms \citep{perozzi2014deepwalk, planetoid2016revisiting}. What sets GNNs apart from previous models is their ability to jointly model both structural information and node attributes. In principle, all GNN models consist of a message passing scheme that propagates feature information of a node to its neighbors. Most GNN architectures use a learnable parameter matrix for projecting features to a different feature space. Usually, two or more of such layers are used along with a nonlinearity (e.g. ReLU). With normalized adjacency matrix $\hat{A} = \tilde{D}^{-1/2} \tilde{A} \tilde{D}^{-1/2}$ a two-layer GCN model \citep{kipf2017gcn} can be expressed as
\begin{equation}
Y_\text{GCN} = \text{softmax}\left( \hat{A}~\text{ReLU} \left(  \hat{A} X \theta^{(0)}\right) \theta^{(1)}\right),
\end{equation}
where $\tilde{A}$ and $\tilde{D}$ are the adjacency matrix and the degree matrix of graph $G$. $\theta^{(0)}$ and $\theta^{(1)}$ are the weight matrices of two neural layers.

\cite{sgc2019} arrived at a much simpler model named SGC by removing hidden layers and nonlinear activations in GCN model. This model can be written as
\begin{equation}
Y_{\text{SGC}} = \text{softmax}\left(\hat{A}^k X \theta \right).
\end{equation}

\subsection{Active Learning}
AL research has contributed a multitude of approaches for training supervised learning models with less labeled data. We recommend \cite{settles2009active} for a detailed review of AL.The objective of most existing AL approaches is to select the most informative instance for labeling. Uncertainty sampling is the most commonly used AL approach. \citet{gal2016dropout} propose using dropout at evaluation time as a way to calculate the model uncertainty of convolutional neural networks (CNN). \citet{gal2017bald} provide a comparison of various acquisition functions for quantifying the model uncertainty of CNN models.  The use of meta-learning for AL has been considered in a few recent works~\citep{woodward2017active, bachman2017learning}. However, these algorithms are designed for the few-shot learning setting and tied to RNN-based meta-learning models such as matching networks~\citep{vinyals2016matching}. Additionally, their reliance on reinforcement learning makes the training difficult. In contrast, our approach is built on model agnostic meta-learning  (MAML)~\citep{finn2017maml} which is efficient and can be used with a variety of supervised loss functions.

%\subsection{Active Learning for Graph Data}
%Compared to applications of AL on image data, only a limited number of AL models have been developed for graph data. Previous work on applying AL on graph data~\citep{gu2012towards, bilgic2010networkdata, ji2012variance} depend on earlier classification models such as Gaussian random fields, in which the features of nodes are not being used. Therefore, selecting query nodes uniformly in random coupled with a  recent graph neural network (GNN) model can easily outperform such AL models. AL models that use recent GNN architectures~\citep{age2017cai, gao2018active} are limited and they rely on linear combinations of uncertainty and various heuristics such as node centrality measures. As we show in this paper, the performance of such models is inconsistent; efficient on some datasets, worse than random sampling on other datasets.

\section{Conclusion}
In this paper we introduced MetAL, a principled approach to perform active learning on graph data. We expressed the semi-supervised active learning problem as a bilevel optimization problem and demonstrated that meta-gradients can be used to make the bilevel optimization problem tractable. Empirical performance on benchmark attributed graphs drawn from multiple domains shows that our proposed method is superior to existing heuristics-based AL algorithms. We further show the importance of performing exploration in addition to exploitation in AL problems. Adaptively learning the exploration coefficient using the feedback from the active learner is an interesting future direction.

In this work, we acquire a single unlabeled instance in each AL step and retrain the classifier. However, acquiring a batch of instances can make the learning process more efficient by reducing the number of retraining steps. We consider adapting MetAL for batch-mode acquisition as another avenue for future improvement. Additionally, understanding which characteristics of an attributed graph make AL easier or difficult is an open research problem. Such an understanding will lead to more efficient AL algorithms in the future.

%\acks{Acknowledgements should go at the end, before appendices and references.}

%\bibliographystyle{plain}
\bibliography{metal2020}

\appendix

%\section{First Appendix}\label{apd:first}
%
%This is the first appendix.
%
%\section{Second Appendix}\label{apd:second}
%
%This is the second appendix.

\end{document}